\documentclass[letterpaper]{article} 
\usepackage{aaai2026}  
\usepackage{times}  
\usepackage{helvet}  
\usepackage{courier}  
\usepackage[hyphens]{url}  
\usepackage{graphicx} 
\usepackage{natbib}  
\usepackage{caption} 
\usepackage{bm}
\usepackage{amssymb}
\usepackage{algorithm}
\usepackage{algorithmic}
\usepackage{booktabs}
\usepackage{multirow}
\usepackage{xcolor}         
\usepackage{amsfonts}
\usepackage{colortbl} 
\usepackage[most]{tcolorbox}
\usepackage{utfsym}
\usepackage{amsthm}
\usepackage{newfloat}
\usepackage{listings}
\DeclareCaptionStyle{ruled}{labelfont=normalfont,labelsep=colon,strut=off} 
\floatstyle{ruled}
\newfloat{listing}{tb}{lst}{}
\floatname{listing}{Listing}
\title{Breaking the Trade-Off Between Faithfulness and Expressiveness for Large Language Models}
\author{
    Chenxu Yang, Qingyi Si, Zheng Lin
}
\affiliations{
   \textsuperscript{\rm 1}Institute of Information Engineering, Chinese Academy of Sciences, Beijing, China \\
  \textsuperscript{\rm 2}School of Cyber Security, University of Chinese Academy of Sciences, Beijing, China \\
  \textsuperscript{\rm 3}Huawei Technologies Co., Ltd.\\
  \textrm{\{}yangchenxu,linzheng\textrm{\}}@iie.ac.cn; 
  {siqingyi@huawei.com}
}

\usepackage{bibentry}

\begin{document}

\maketitle

\begin{abstract}
Grounding responses in external knowledge represents an effective strategy for mitigating hallucinations in Large Language Models (LLMs). However, current LLMs struggle to seamlessly integrate knowledge while simultaneously maintaining \textit{faithfulness} (or \textit{fidelity}) and \textit{expressiveness}, capabilities that humans naturally possess. This limitation results in outputs that either lack support from external knowledge, thereby compromising faithfulness, or appear overly verbose and unnatural, thus sacrificing expressiveness.
In this work, to break the trade-off between faithfulness and expressiveness, we propose \textbf{Co}llaborative \textbf{De}coding (\textbf{CoDe}), a novel approach that dynamically integrates output probabilities generated with and without external knowledge. This integration is guided by distribution divergence and model confidence, enabling the selective activation of relevant and reliable expressions from the model's internal parameters. Furthermore, we introduce a knowledge-aware reranking mechanism that prevents over-reliance on prior parametric knowledge while ensuring proper utilization of provided external information.
Through comprehensive experiments, our plug-and-play CoDe framework demonstrates superior performance in enhancing faithfulness without compromising expressiveness across diverse LLMs and evaluation metrics, validating both its effectiveness and generalizability.

\end{abstract}


\section{Introduction}
\begin{figure}[!t]
  \centerline{\includegraphics[scale=0.24]{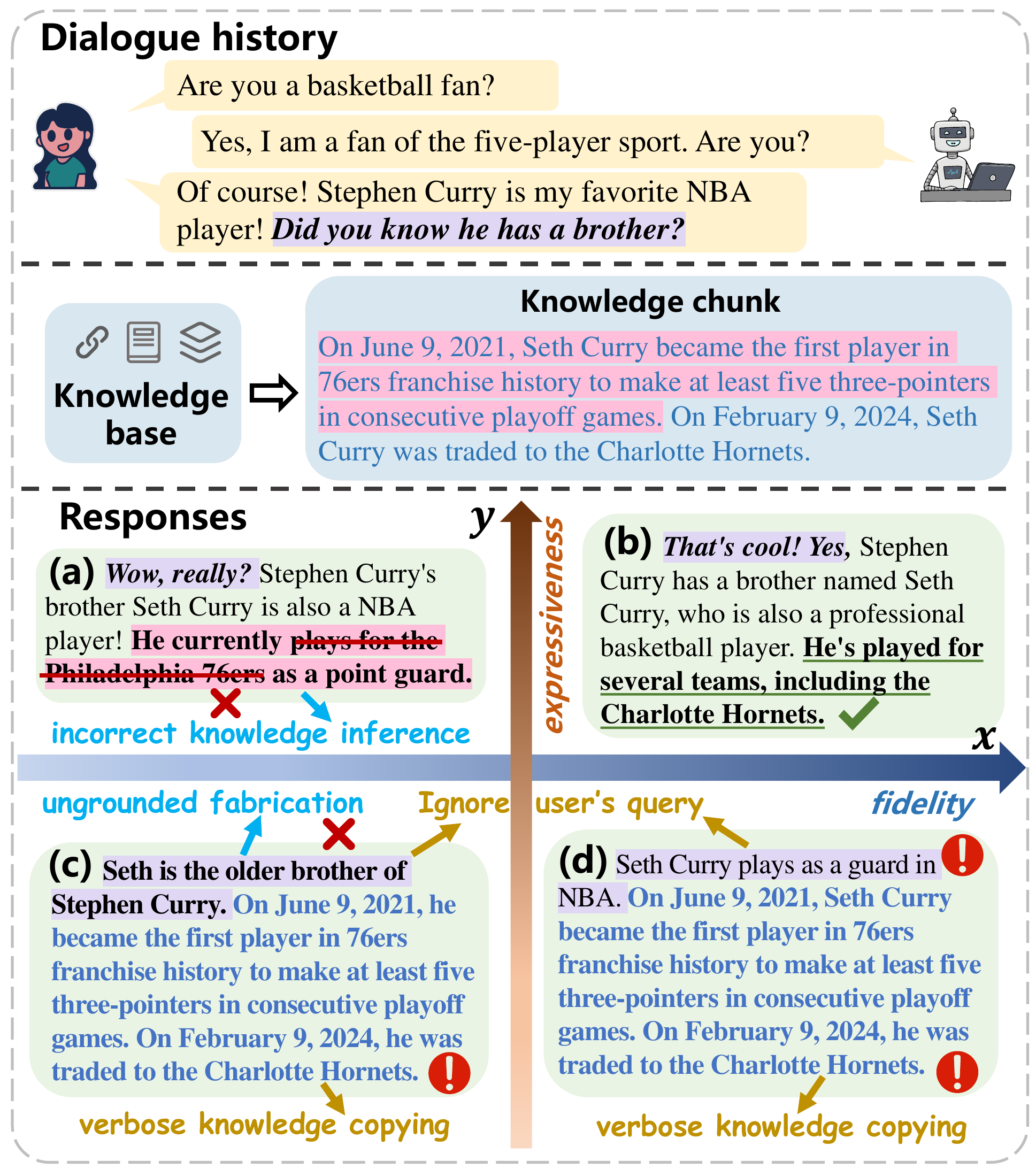}}
  \caption{Examples exhibits the trade-off between expressiveness and faithfulness in LLMs. Higher x-coordinates correspond to higher faithfulness, and higher y-coordinates correspond to better expressiveness. Examples (a), (c), and (d) are constrained by the trade-off, whereas our approach break it and generate responses like (b).}
  \label{fig-example}
\end{figure}

Although large language models (LLMs) have demonstrated remarkable performance across diverse tasks in recent studies \cite{bai2023qwen,yang2023baichuan,touvron2023llama2,chatgpt,openai2023gpt4,yang2024orthogonalfinetuningdirectpreference,dai2025sgrpoearlyexitreinforcement}, they remain susceptible to hallucination, producing content that appears plausible yet lacks factual accuracy \cite{Ji_2023, 10.1145/3703155survey}. Research indicates that this phenomenon arises from fundamental limitations in LLMs, including constrained knowledge boundaries \cite{Ren2023InvestigatingTF}, insufficient coverage of long-tail knowledge \cite{10.5555/3618408.3619049longtai}, and outdated parametric knowledge. These inherent constraints significantly hinder the practical deployment of LLMs.
To address these challenges, augmenting LLMs with external knowledge through incorporation into model inputs has emerged as a promising solution, demonstrating substantial improvements in the factual accuracy of generated content. The Retrieval-Augmented Generation (RAG) paradigm, in particular, has gained widespread adoption as an effective approach to this problem.

However, external-knowledge-augmented LLMs, such as those employing RAG, continue to face two fundamental challenges. First, they frequently generate content that contradicts or lacks support from provided knowledge, as shown in responses (a) and (c) of Figure \ref{fig-example}. Second, they struggle to integrate external knowledge naturally, often producing responses with poor interactivity, dullness, and redundancy \cite{chen-etal-2023-fidelity,yang-etal-2023-multi-level}. Response (d) in Figure 1 illustrates this limitation: the model merely echoes the provided knowledge without addressing the user's greeting, substantially diminishing conversational engagement.
While existing methods address the first challenge \cite{deng-etal-2023-towards-huyue,10.1609/aaai.v37i11.26596sdu,zhang-etal-2024-selfalign,liang-etal-2024-learning}, they neglect or even worsen the second. An effective LLM must balance two requirements: it must generate responses grounded in the provided knowledge, a property we define as faithfulness (or fidelity), and it must leverage external knowledge creatively to produce natural, diverse, and engaging responses, which we term expressiveness. We provide detailed definitions for these two properties in Section 3.2.
Previous work by \citet{chawla-etal-2024-investigating} identified this fidelity-expressiveness conflict through input masking experiments, yet failed to propose a practical solution.
We extend this analysis to decoding strategies in Section 3.3, revealing that deterministic decoding sacrifices expressiveness while stochastic decoding compromises fidelity. This comprehensive understanding enables our principled solution.

To resolve this trade-off in LLMs, we propose Collaborative Decoding (CoDe), a novel method that dynamically elicits relevant and factual natural expressions from the model's internal parameters. CoDe achieves this by integrating output probability distributions generated with and without external knowledge, guided by their distributional divergence and model confidence.
Specifically, we employ Jensen-Shannon Divergence (JSD) to quantify the distribution differences and combine local confidence (maximum probability) with global uncertainty (entropy) to measure model confidence, facilitating complementary cooperation between two distributions.
By enhancing expressiveness without introducing stochasticity, our approach effectively circumvents hallucinations typically associated with sampling-based methods.
Additionally, we introduce a knowledge-aware reranking mechanism to prevent overreliance on parametric knowledge at the expense of external knowledge. This mechanism reranks the top-k candidate tokens based on their alignment with external knowledge, evaluated by both semantic similarity and attention patterns, thereby ensuring faithfulness to the provided knowledge.


Our contributions are summarized as follows:
 \begin{itemize}
    \item We investigate the trade-off between expressiveness and faithfulness in external-knowledge-augmented LLMs, focusing on decoding strategies.
    \item We introduce CoDe, a novel method that simultaneously enhances both faithfulness and expressiveness in knowledge-grounded scenarios without requiring additional training, model, or generation budgets.
	\item We demonstrate CoDe's effectiveness and generalizability through comprehensive experiments, comparing against ten baseline decoding methods across six LLMs, three datasets, and nine evaluation metrics.
\end{itemize}

 \section{Related Work}



\subsection{Hallucinations in Text Generation}
Hallucination refers to the generation of LLMs appears plausible but is factually incorrect \cite{zhang2023sirens}. The research community has extensively investigated this phenomenon from multiple perspectives, including its underlying causes \cite{mckenna2023sources,dziri2022origin}, detection methodologies \cite{zhang2023enhancing,manakul2023selfcheckgpt,fadeeva2023lmpolygraph}, and mitigation strategies \cite{choi2023kcts,chuang2023dola,li2023inferencetime,yang2025testtimepromptintervention}.
Retrieval-Augmented Generation (RAG) has emerged as a prominent approach for mitigating hallucinations by incorporating external knowledge. Several studies have pursued training-based solutions, constructing preference-aligned or human-annotated datasets to fine-tune models for improved fidelity \cite{liang-etal-2024-learning, zhang-etal-2024-selfalign}. Others have adopted Chain-of-Thought approaches \cite{10.5555/3600270.3602070COT}, externalizing implicit knowledge from the backbone LLM \cite{zhou-etal-2022-think, chae-etal-2023-dialogueDOCTOR, yu-etal-2024-llms} or employing self-reflection mechanisms \cite{asai2024selfrag}. In contrast, our CoDe method offers a lightweight solution that effectively mitigates hallucinations without requiring training, auxiliary models, or time-intensive reflections.




\subsection{Generation Decoding Strategy}

Decoding strategies determine next-token selection from vocabulary probability distributions, including greedy decoding, beam search, and top-k sampling. Nucleus sampling \cite{holtzman2020curious} dynamically selects tokens until reaching a cumulative probability threshold. While stochastic methods enhance diversity, they compromise semantic consistency \cite{Basu2021MirostatAN, su2022contrastive} and increase hallucination rates \cite{dziri-etal-2022-faithdial}.
Recent contrastive decoding methods have recently gained significant attention. Contrastive Decoding \cite{li-etal-2023-contrastive} maximizes expert-amateur log-probability differences for improved fluency. DoLa \cite{chuang2023dola} contrasts mature and pre-mature layer logits to reduce hallucinations. CAD \cite{shi-etal-2024-trustingCAD} amplifies probability differences between context-aware and context-free outputs. VCD \cite{Leng2023MitigatingOHVCD} contrasts original and distorted visual inputs in vision-language models.
Unlike these error-filtering approaches, CoDe employs dynamic collaboration between distributions, simultaneously optimizing both fidelity and expressiveness rather than addressing single limitations.



\section{Preliminaries}

\subsection{Task Formulation}

We consider an LLM parameterized by $\theta$. The model input comprises four components: a task-specific instruction $\mathcal{I}$, multi-turn dialogue history $\bm{h}$, the current user utterance $\bm{u}$, and relevant external knowledge $\bm{k}=(k_1,\dots,k_{m})$ containing $m$ tokens. For notational convenience, we define the conversation context as $\bm{x}= [\mathcal{I}; \bm{h}; \bm{u}]$.

At each time step, the LLM generates the next token based on the input and previously generated tokens $\bm{y}_{< t}$, producing vocabulary logits:
\begin{equation}
\text{logit}_{\theta}(y_t|\cdot)=\mathcal{LLM}_{\theta}(\bm{x}, \bm{k}, \bm{y}_{< t}).
\end{equation}
The probability distribution is obtained via softmax transformation, and various decoding strategies select the next token $y_t$ from the resulting distribution:
\begin{equation}
y\sim p_{\theta}(y_t|\bm{x}, \bm{k}, \bm{y}_{<t})\propto\text{exp}\,\text{logit}_{\theta}(y_t|\cdot).
\end{equation}

\begin{figure}[]
  \centerline{\includegraphics[scale=0.26]{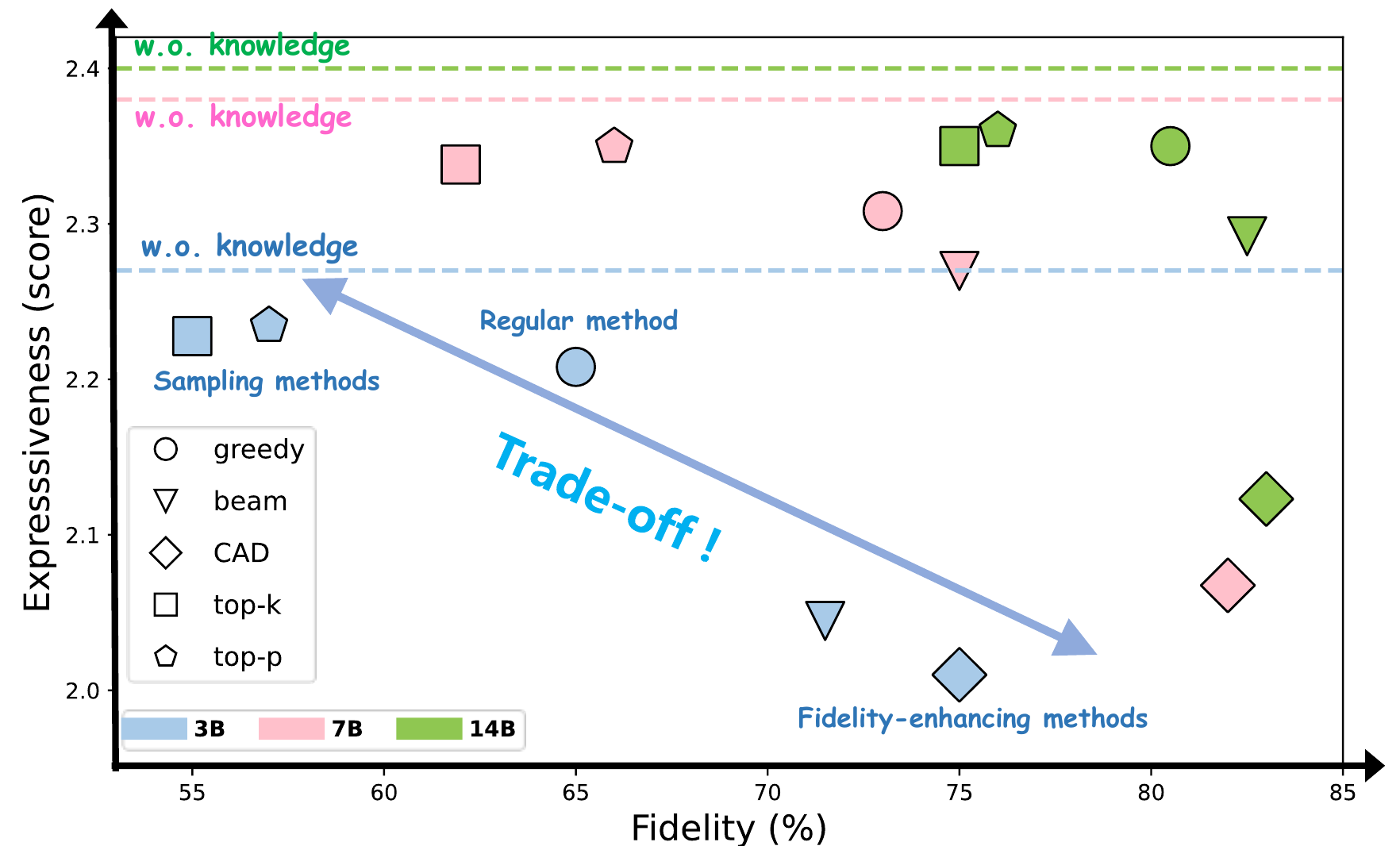}}
  \caption{The trade-off between fidelity and expressiveness of current decoding strategies on Qwen2.5-chat models at different scales. The dashed line indicates the expressiveness score without referring to knowledge.}
  \label{fig-trade-off}
\end{figure}

\subsection{Conceptual Definitions}
\textbf{Faithfulness} (or \textbf{fidelity})  denotes the consistency between generated responses and external knowledge without contradictions. A formal definition and distinction from \textbf{factuality} are provided in Appendix H.

\noindent\textbf{Expressiveness} encompasses three key dimensions:
(1) \textit{context-aware interaction}, prioritizing conversational coherence and user engagement over mere information delivery; (2) \textit{natural knowledge integration}, extracting and seamlessly incorporating relevant information rather than copying source text; and (3) \textit{linguistic diversity}, exhibiting varied expression patterns while avoiding formulaic language.
\subsection{Pilot Observations and Insights} \label{pilot3.2}

\paragraph{There remains considerable potential for improvement in expressiveness and fidelity.} Integrating external knowledge into LLMs creates a fundamental tension: while improving informativeness, it often diminishes response expressiveness. As shown in Figure \ref{fig-example} (panels c-d) and quantified in Figure \ref{fig-trade-off}, LLMs tend to directly copy external knowledge rather than seamlessly incorporating it, resulting in decreased expressiveness scores. This suggests that LLMs sacrifice discourse coherence and naturalness when prioritizing faithful information transmission.
Moreover, a substantial fidelity gap exists between LLM and human-generated content. Despite external knowledge access, LLMs frequently produce contradictory outputs due to flawed reasoning or conflicts with their parametric knowledge, as illustrated in Figure \ref{fig-example} (panels a, c). Figure \ref{pilot-faith} confirms that even advanced open-source LLMs significantly underperform humans in maintaining knowledge fidelity, highlighting persistent challenges in neural knowledge grounding.

\noindent\textbf{Current decoding strategies reveal a fundamental trade-off between expressiveness and knowledge fidelity.} As illustrated in Figure \ref{fig-trade-off}, this dilemma manifests distinctly across different decoding approaches: deterministic decoding yields content with high fidelity but compromised expressiveness, whereas stochastic decoding enhances linguistic diversity at the cost of factual accuracy. Notably, this trade-off is particularly pronounced in smaller-scale models, which exhibit greater sensitivity to the choice of decoding strategy. This paper aims to break the trade-off by proposing a novel approach that achieve a win-win situation for both faithfulness and expressiveness. The experimental setups for the pilot study are detailed in Appendix A.

\section{Approach}

\begin{figure*}[htbp]
  \centerline{\includegraphics[scale=0.202]{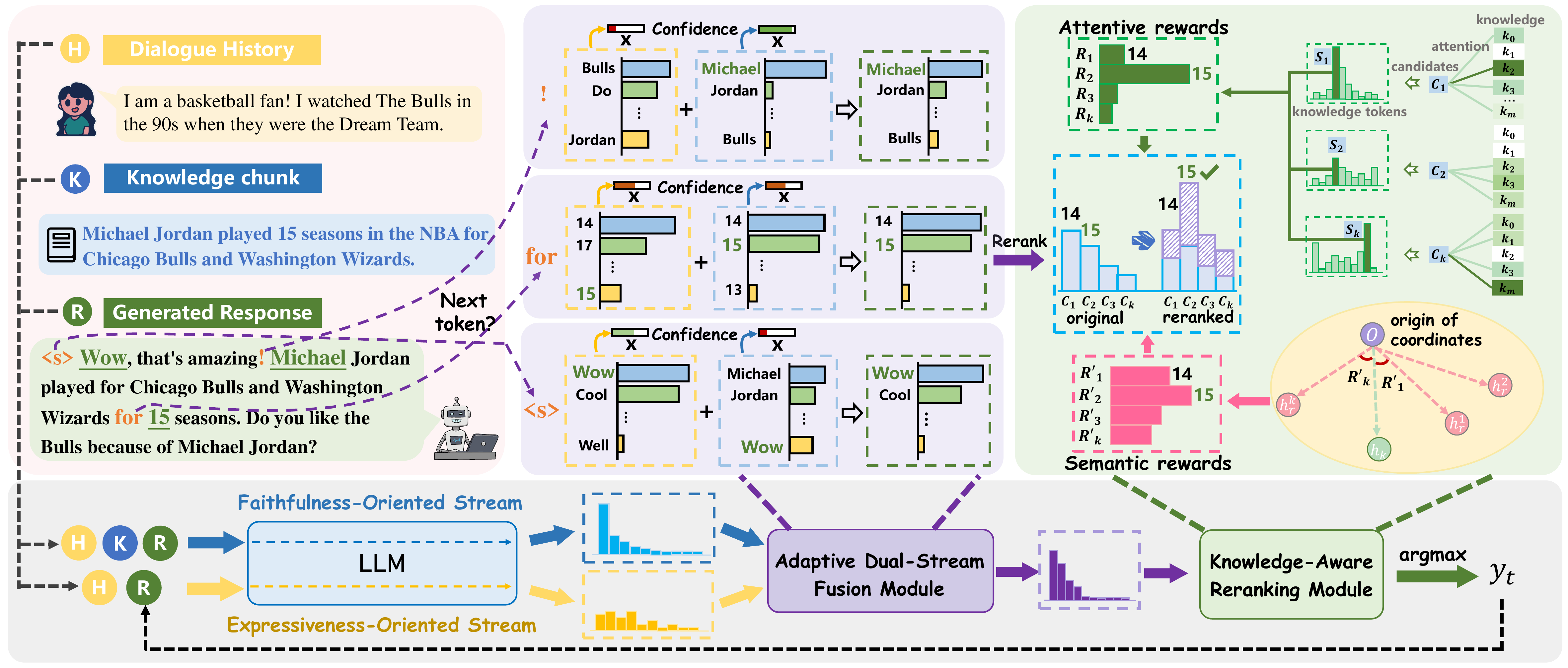}}
  \caption{An overview of the CoDe method, which comprises two key components: (1) an Adaptive Dual-Stream Fusion Module that dynamically integrates internal and external knowledge by leveraging model confidence and distribution divergence, and (2) a Knowledge-Aware Reranking Module that employs semantic and attentive rewards to select faithful tokens.}
  \label{figmethod}
\end{figure*}

This section presents Collaborative Decoding (CoDe), a novel method for external-knowledge-augmented LLMs comprising two key components, as illustrated in Figure \ref{figmethod}.


\subsection{Adaptive Dual-Stream Fusion} \label{subsection3.1}


As shown in Section 3.3, models with external knowledge input tend to copy knowledge fragments, thereby diminishing expressiveness. While stochastic decoding methods like top-k \cite{fan-etal-2018-topk} and nucleus sampling \cite{holtzman2020curious} mitigates this issue, their probabilistic nature inevitably induces hallucinations. We hope to propose a deterministic approach that enhances expressiveness without sacrificing factual accuracy.


\paragraph{Distribution Collaboration.} Inspired by contrastive decoding \cite{li-etal-2023-contrastive}, we propose a dual-stream fusion approach that emphasizes \textit{complementary collaboration} rather than error filtering through contrast. CoDe generates two output distributions: an expressiveness-oriented stream conditioned solely on conversation context $x$, and a faithfulness-oriented stream conditioned on both context $x$ and external knowledge $k$. These streams are then fused to create a collaborative distribution that breaks the trade-off between expressiveness and faithfulness :
\begin{equation}
\begin{aligned}
p_{CoDe}(y_t) = & \;\text{softmax}  [ \alpha \, \text{logit}_\theta(y_t |  \bm{x},\bm{k}, \bm{y}_{<t}) \\
 &+ (1- \alpha)\, \text{logit}_\theta(y_t | \bm{x}, \bm{y}_{<t}) ],
\label{eq:code}
\end{aligned}
\end{equation}
where larger $\alpha$ indicates more weight on the faithfulness-oriented stream.
The Equation \ref{eq:code} could also be written as:
\begin{equation}
\begin{aligned}
p_{CoDe} \propto  p_\theta(y_t |  \bm{x}, \bm{y}_{<t}) \left( \frac{p_\theta(y_t |   \bm{x},\bm{k}, \bm{y}_{<t})}{p_\theta(y_t | \bm{x}, \bm{y}_{<t})} \right)^{\alpha}.
\label{eq: 2}
\end{aligned}
\end{equation}



In this formulation, $p_\theta(y_t | \bm{x}, \bm{y}_{<t})$ represents the prior distribution based solely on the model's parametric knowledge, while $p_\theta(y_t | \bm{x},\bm{k}, \bm{y}_{<t})$ denotes the posterior distribution conditioned on external knowledge $\bm{k}$. CoDe leverages pointwise mutual information (PMI) between $\bm{k}$ and $y_t$ to dynamically recalibrate output probabilities, amplifying tokens strongly associated with external knowledge.

\paragraph{Adaptive Fusion Weights $\alpha$.} To prevent hallucinations from low-probability tokens, we adaptively modulate $\alpha$ based on \textbf{model confidence} and \textbf{distribution divergence}. When internal knowledge exhibits low relevance or high uncertainty, CoDe reduces parametric reliance and prioritizes external knowledge integration.
\begin{equation}
\alpha = \frac{{\delta}\cdot{\mathcal C^k_t}}{ {\mathcal C^c_t} + {\delta}\cdot{\mathcal C^k_t}},
\label{eq: m0}
\end{equation}
where $\mathcal C^k_t$ and $\mathcal C^c_t$ denotes the confidence of posterior and prior knowledge, $\delta$ denotes the distribution divergence.

Recent work on LLM hallucination determine when to trust LLMs based on uncertainty \cite{manakul2023selfcheckgpt,huang2023look,duan2023shifting}, we adopt the uncertainty-based confidence framework of \citet{zhang-etal-2023-enhancing-uncertainty}, quantifying factual confidence through local confidence $p_{max}$ (maximum token probability) and global uncertainty $\mathcal{H}_t$ (distribution entropy):
\begin{equation}
\begin{aligned}
&p_{max}=\max_{y_t\in\mathcal{V}} \,p(y_t), \\
\mathcal{H}_t=-&\sum_{y_t\in\mathcal{V}}p(y_t)*\log_2(p(y_t)).
\label{eq: 7}
\end{aligned}
\end{equation}


 
We then synthesize $p_{max}$ and $\mathcal{H}_t$ using the geometric mean function, deriving the confidence score $\mathcal C_t$ as follows:
\begin{equation}
\mathcal C_{t}= \sqrt[2]{\frac{ p_{max}}{\mathcal{H}_t + \eta}  },
\label{eq: m1}
\end{equation}
where $\eta$ is a small constant prevents value overflow.


When the prior and posterior distributions diverge significantly, this signals a conflict between internal and external knowledge, prompting us to reduce the prior weight and prioritize external information. Conversely, when the distributions align closely, indicating consistent knowledge representations, we increase the prior weight to leverage pre-trained knowledge for enhanced expressiveness. To implement this adaptive mechanism, we introduce a dynamic parameter $\delta$ in the design of $\alpha$:
\begin{equation}
\delta = \gamma \cdot \exp(\text{JSD}\left(p_c(y_t) \| p_k(y_t)\right)),
\label{eq: m2}
\end{equation}
where $\text{JSD}(\cdot,\cdot)$ denotes the Jensen-Shannon Divergence, and $\gamma$ is a scale factor.

\begin{table*}[!h]
\centering
\setlength{\tabcolsep}{5pt} 
\renewcommand{\arraystretch}{0.95} 
\scalebox{0.9}{
\begin{tabular}{lccccccccccccccc}
\toprule
  \multirow{3}{*}{Method} & \multicolumn{7}{c}{\textbf{\textsc{FaithDial}}} & \multicolumn{7}{c}{\textbf{HalluDial}} \\ 
 \cmidrule(lr){2-8} \cmidrule(lr){9-15} 
 & \multicolumn{3}{c}{\textbf{Expressiveness}} & \multicolumn{3}{c}{\textbf{Faithfulness}} & \multirow{2}{*}{\textbf{Avg.}} & \multicolumn{3}{c}{\textbf{Expressiveness}} & \multicolumn{3}{c}{\textbf{Faithfulness}} & \multirow{2}{*}{\textbf{Avg.}} \\ 
 \cmidrule(lr){2-4}  \cmidrule(lr){5-7} \cmidrule(lr){9-11}  \cmidrule(lr){12-14}
 & \textbf{DIV} & \textbf{COH} & \textbf{CRE} & \multicolumn{1}{l}{\textbf{F-Critic}} & \multicolumn{1}{l}{\textbf{H-Judge}} & \multicolumn{1}{l}{\textbf{K-BP}} & & \textbf{DIV} & \textbf{COH} & \textbf{CRE} & \multicolumn{1}{l}{\textbf{F-Critic}} & \multicolumn{1}{l}{\textbf{H-Judge}} & \multicolumn{1}{l}{\textbf{K-BP}} & \\ \hline
Greedy & 31.4 & \underline{57.3} & 30.0 & 28.5 & 86.9 & 60.9 & 49.2
& 36.9 & \underline{64.6} & 30.1 & 30.2 & 87.8 & 61.2 & 51.8 \\
Beam & 30.8 & \textbf{57.6} & 25.6 & 31.3 & 89.3 & \underline{64.9} & 49.9
& 36.1 & 64.4 & 24.3 & 31.5 & 89.0 & 65.7 & 51.8 \\
CS & 33.9 & 55.4 & 30.0 & 30.9 & 83.2 & 58.7 & 48.7
& 37.5 & \underline{64.6} & 30.2 & 30.4 & 88.7 & 60.9 & 52.0 \\
FECS & 32.8 & 56.8 & 28.0 & 31.6 & 88.1 & 63.5 & 50.1
& 39.4 & 64.4 & 30.4 & 31.0 & 89.9 & 64.3 & \underline{53.2} \\ 
top-k & \textbf{36.2} & 57.2 & \textbf{34.5} & 21.8 & 75.3 & 56.8 & 47.0
& \underline{40.7} & 63.8 & \textbf{36.0} & 21.1 & 73.0 & 56.4 & 48.5 \\
Nucleus & \underline{35.6} & 57.2 & \underline{34.3} & 23.4 & 79.7 & 57.4 & 47.9
& 39.9 & 64.1 & \underline{34.6} & 25.3 & 79.0 & 57.8 & 50.1 \\
F-Nucleus & 34.1 & \underline{57.3} & 32.9 & 24.3 & 82.0 & 58.6 & 48.2
& 38.5 & 64.4 & 32.3 & 25.7 & 82.4 & 59.1 & 50.4 \\ 
CD & 35.0 & 55.9 & 31.3 & 22.6 & 76.2 & 57.0 & 46.3
& 38.4 & 62.9 & 30.5 & 24.1 & 78.9 & 57.3 & 48.7 \\
DoLa & 32.8 & 56.2 & 32.3 & 31.4 & 87.3 & 61.2 & \underline{50.2}
& 39.0 & 64.0 & 33.6 & 32.2 & 89.1 & 60.4 & 53.0 \\
CAD & 29.2 & 52.8 & 21.7 & \underline{32.1} & \underline{90.4} & \textbf{67.0} & 48.9
& 35.4 & 59.8 & 22.3 & \underline{33.6} & \textbf{90.4} & \underline{67.3} & 51.5 \\ 
\textbf{CoDe} & \underline{35.6} & \textbf{57.6} & 29.9 & \textbf{32.4} & \textbf{90.8} & \textbf{67.0} & \textbf{52.2}
& \textbf{40.9} & \textbf{64.9} & 29.8 & \textbf{34.3} & \textbf{90.4} & \textbf{67.5} & \textbf{54.6} \\ 
\bottomrule
\end{tabular}
}
\caption{Automatic evaluation results on the {\textsc{FaithDial}} and {HalluDial} dataset (Llama2-7B-chat). The best results are highlighted with \textbf{bold}. The second-best results are highlighted with \underline{underline}. Avg. denotes the average across all metrics.}
\label{tab1}
\end{table*}

\subsection{Knowledge-Aware Reranking} \label{subsection3.2}


To prevent the model from being overly confident in its prior parameter knowledge and thereby ignoring external knowledge, we introduce a knowledge-aware reranking mechanism that further refines CoDe's output distribution:
\begin{equation}
\begin{aligned}
 &\hat{p}_{CoDe}(y_t) =\; \text{topK}\Big\{(1-\beta)\, {p}_{CoDe}(y_t) + \\
                  \frac{\beta}{2}& \Big[ { \max_{k_i \in \bm{k}}  \{ \text{sim}(h_{y_{t}},h_{k_i})\} } 
+ {\max_{k_j \in \bm{k}}\{\text{att}(y_t,{k_j})\}} \Big]
                \Big\},
\label{eq: 3}
\end{aligned}
\end{equation}
where $\beta$ controls the fidelity amplification strength, $h$ represents hidden states, $\text{sim}(\cdot,\cdot)$ denotes cosine similarity, and $\text{att}(y_t,k_j)$ represents the max-pooled attention weight between token $y_t$ and knowledge element $k_j$ across all layers and heads.
The knowledge-aware reranking mechanism ensures fidelity through two complementary rewards: (1) semantic reward, which favors tokens with high cosine similarity to external knowledge tokens, and (2) attentive reward, which prioritizes tokens exhibiting stronger attention to knowledge segments. As illustrated in Figure \ref{figmethod}, when internal and external knowledge conflict (e.g., the model's "14 seasons" versus the correct "15 seasons" for Jordan), this mechanism amplifies external knowledge awareness, enabling accurate token selection ($14\rightarrow 15$).

The final token $y_t$ is selected from the top-K candidates based on the combined score of fidelity and expressiveness:
\begin{equation}
y_t = \arg\max\hat{p}_{CoDe}(y_t).
\end{equation}

\section{Experiments}

\subsection{Experimental Setup}

\textbf{Datasets and Models.}
e evaluated CoDe on three information-seeking dialogue datasets—\textsc{FaithDial} \cite{dziri-etal-2022-faithdial}, HalluDial \cite{luo2024halludiallargescalebenchmarkautomatic}, and WoW \cite{WoW}—which provide dialogue contexts with external knowledge for response generation. Additionally, we tested on three non-conversational benchmarks: Natural Questions \cite{kwiatkowski-etal-2019-natural-NQ}, NQ-SWAP \cite{longpre2022entitybasedknowledgeconflictsquestionNQswap}, and HalluEval \cite{li2023halueval}, demonstrating CoDe's effectiveness in faithfulness-only scenarios. 
We evaluated six LLMs across different scales and architectures: Llama2-7B-chat \cite{touvron2023llama2}, Llama-3.1-8B-chat \cite{grattafiori2024llama3herdmodels}, Mistral-7B-Instruct-v0.2 \cite{jiang2023mistral7b}, and Qwen-2.5 series (3B, 7B, 14B) \cite{qwen2025qwen25technicalreport}. Dataset and implementation details are in Appendices B and D.

\begin{figure}[!t]
  \centerline{\includegraphics[scale=0.25]{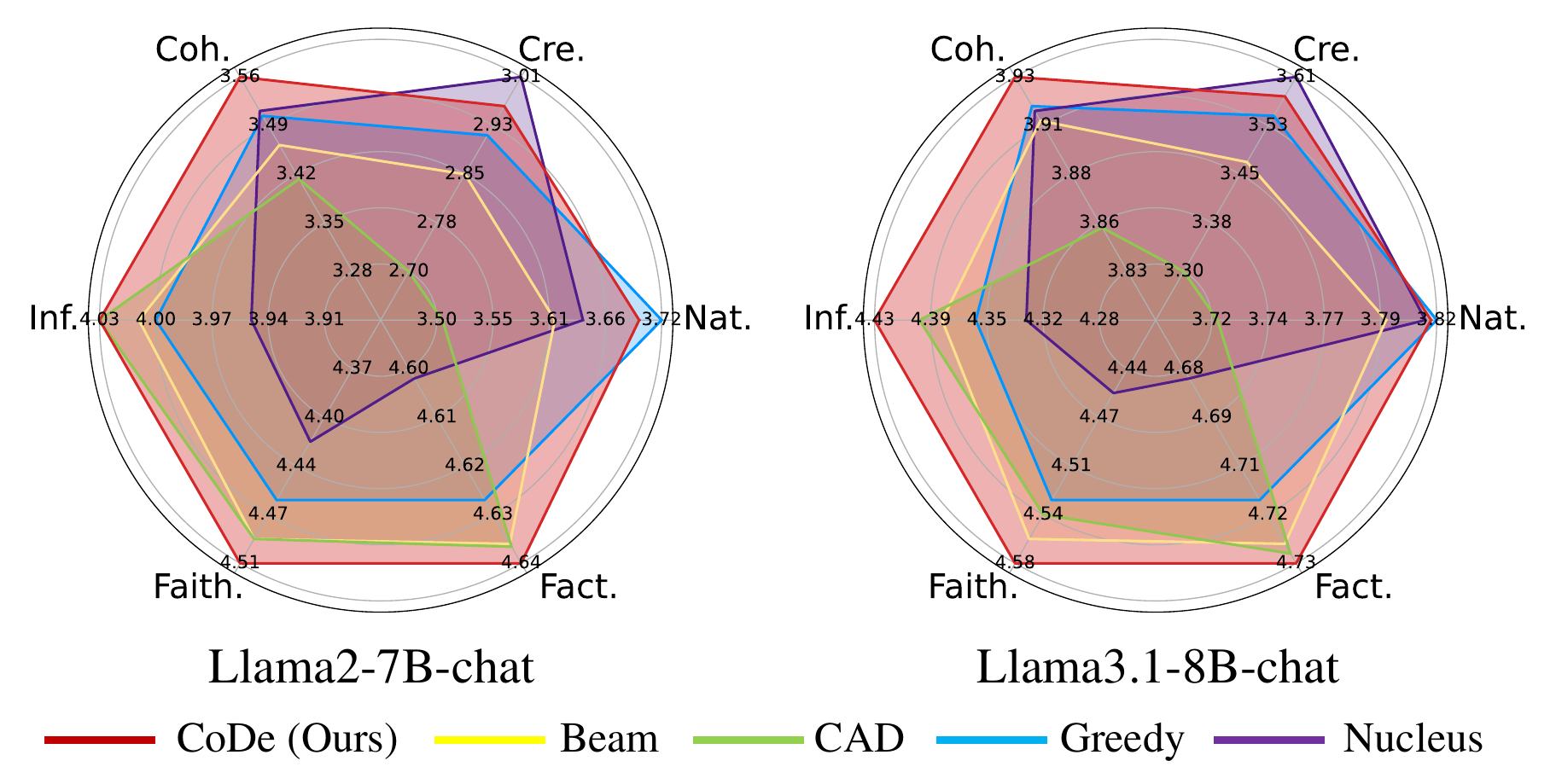}}
  \caption{LLM-based evaluation results on the \textsc{FaithDial} dataset (Llama2-7B-chat).}
  \label{g4-eval}
\end{figure} 
\noindent \textbf{Baselines.}
We choose ten decoding methods as the baselines. 
\noindent\textbf{Search Methods:} Greedy Decoding (Greedy), Beam Search (Beam), Contrastive Search (CS) \cite{su2022contrastive}, and FECS \cite{chen-etal-2023-fidelity}. 
\noindent\textbf{Stochastic Methods:} Top-k Sampling \cite{fan-etal-2018-topk}, Nucleus Sampling (Nulceus) \cite{holtzman2020curious}, and Factual-Nucleus Sampling (F-Nucleus) \cite{lee2023factuality}. 
\noindent\textbf{Contrastive Methods:}  Contrastive Decoding (CD) \cite{li-etal-2023-contrastive}, DoLa \cite{chuang2023dola}, and Context-Aware Decoding (CAD) \cite{shi-etal-2024-trustingCAD}. The details of the baseline introduction and hyperparameter settings are found in the Appendix \ref{baseline}.

\subsection{Experimental Results}

\begin{figure}[!t]
  \centerline{\includegraphics[scale=0.3]{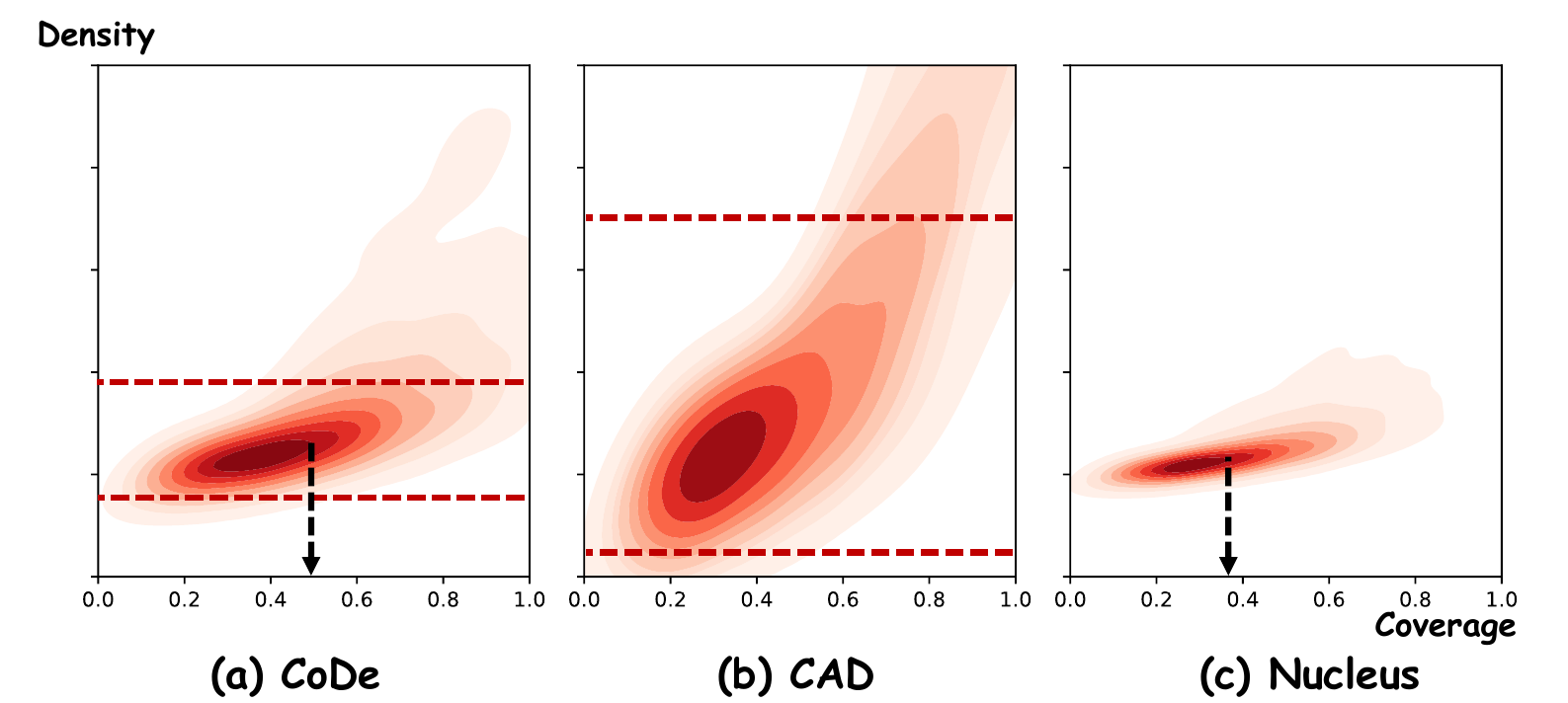}}  \caption{Knowledge utilization patterns across CoDe, CAD, and Nucleus decoding methods. Bottom-right concentration indicates superior performance.}
  \label{density}
\end{figure}

\begin{table*}[]
\centering
\setlength{\tabcolsep}{4.5pt}
\scalebox{0.85}{
\begin{tabular}{llcccccccccc}
\toprule
\multirow{2}{*}{Model} & \multirow{2}{*}{\textbf{Method}} & \multicolumn{3}{c}{\textbf{Expressiveness}} & \multicolumn{3}{c}{\textbf{Faithfulness}} & \multicolumn{3}{c}{\textbf{Quality}} & \multirow{2}{*}{\textbf{Avg.}} \\ 
\cmidrule(lr){3-5} \cmidrule(lr){6-8} \cmidrule(lr){9-11} 
 &  & \textbf{DIV} & \textbf{COH} & \textbf{CRE} & \multicolumn{1}{l}{\textbf{F-Critic}} & \multicolumn{1}{l}{\textbf{H-Judge}} & \multicolumn{1}{l}{\textbf{K-BP}} & \multicolumn{1}{l}{\textbf{BLEU-2/4}} & \multicolumn{1}{l}{\textbf{METEOR}} & \multicolumn{1}{l}{\textbf{ROUGE-L}} & \\ \hline
 \multirow{4}{*}{Mistral-7B-Instruct-v0.2} & \textbf{greedy} & 34.2 & 59.5 & 41.3 & 21.8 & 89.8 & 59.9 & 15.0/6.7 & 19.6 & 25.2 & 37.3 \\
& \textbf{top-k} & {35.2} & 59.5 & \textbf{46.8} & 16.8 & 87.0 & 57.4 & 14.3/6.1 & 18.9 & 23.7 & 36.6 \\
  & \textbf{CAD} & 33.1 & 58.3 & 35.2 & 23.9 & 91.2 & 62.5 & 15.0/6.6 & 20.4 & \textbf{25.3} & 37.1 \\
 & \textbf{CoDe} & \textbf{35.4} & \textbf{59.9} & 38.7 & \textbf{24.3} & \textbf{91.3} & \textbf{62.7} & \textbf{15.5/6.9} & \textbf{20.6} & 25.0 & \textbf{38.0} \\ \hline
  \multirow{4}{*}{Llama-3.1-8B-chat} & \textbf{greedy} & 34.4 & 53.4 & 34.2 & 46.6 & 92.2 & 67.7 & 21.6/10.4 & 22.2 & \textbf{31.0} & 41.4 \\
  & \textbf{top-k} & \textbf{36.0} & 53.2 & \textbf{35.1} & 42.9 & 91.5 & 62.9 & 19.8/9.4 & 20.6 & 28.8 & 40.0 \\
  & \textbf{CAD} & 29.7 & 50.5 & 23.8 & 49.7 & 93.3 & \textbf{72.5} & 21.0/9.8 & 23.0 & 30.3 & 40.4 \\
 & \textbf{CoDe} & {35.7} & \textbf{54.5} & 33.8 & \textbf{50.2} & \textbf{94.0} & 71.7 & \textbf{21.9/10.8} & \textbf{23.5} & \textbf{30.8} & \textbf{42.7} \\ \hline
\multirow{4}{*}{Qwen-2.5-3B-chat} 
& \textbf{greedy} & 37.7 & 52.4 & 37.6 & 38.7 & 90.5 & 56.2 & 18.6/8.5 & 16.2 & 25.8 & 38.2 \\
& \textbf{top-k} & \textbf{40.0} & 50.7 & \textbf{45.9} & 29.0 & 85.4 & 53.8 & 16.9/7.5 & 15.5 & 23.9 & 36.9 \\
  & \textbf{CAD} & 34.8 & 48.5 & 31.9 & 39.6 & 91.4 & 61.0 & 19.1/8.6 & 17.3 & 26.5 & 37.9 \\
 & \textbf{CoDe} & 39.4 & \textbf{54.4} & 37.0 & \textbf{42.9} & \textbf{92.9} & \textbf{61.3} & \textbf{20.9/9.8} & \textbf{18.9} & \textbf{27.4} & \textbf{40.5} \\ \hline
\multirow{4}{*}{Qwen-2.5-7B-chat} 
& \textbf{greedy} & 36.8 & 55.7 & 40.5 & 36.0 & 91.2 & 61.6 & 16.8/7.6 & 18.5 & 25.5 & 39.0 \\
& \textbf{top-k} & 37.3 & 54.8 & \textbf{45.5} & 32.8 & 88.7 & 58.0 & 15.2/7.0 & 17.6 & 24.4 & 38.1 \\
  & \textbf{CAD} & 35.2 & 52.6 & 34.6 & 38.4 & \textbf{92.8} & 63.6 & \textbf{17.8/8.0} & 20.5 & 26.1 & 39.0 \\
 & \textbf{CoDe} & \textbf{37.7} & \textbf{55.8} & 40.8 & \textbf{39.6} & \textbf{92.8} & \textbf{64.8} & 17.6/\textbf{8.0} & \textbf{20.6} & \textbf{26.4} & \textbf{40.4} \\ \hline
\multirow{4}{*}{Qwen-2.5-14B-chat} 
& \textbf{greedy} & 37.8 & \textbf{53.6} & 39.3 & 36.6 & 91.9 & 65.4 & 21.7/10.3 & 21.6 & 30.1 & 40.8 \\
& \textbf{top-k} & 38.5 & 53.4 & \textbf{43.8} & 36.2 & 91.6 & 63.8 & 21.3/10.0 & 21.5 & 29.4 & 41.0 \\
  & \textbf{CAD} & 35.1 & 52.8 & 36.5 & 36.4 & 91.9 & \textbf{66.5} & 22.0/10.3 & 21.4 & 30.3 & 40.3 \\
 & \textbf{CoDe} & \textbf{38.6} & \textbf{53.6} & 39.6 & \textbf{36.9} & \textbf{92.6} & 66.2 & \textbf{22.4/10.5} & \textbf{22.0} & \textbf{30.7} & \textbf{41.3} \\ 
\bottomrule
\end{tabular}
}
\caption{Automatic evaluation results compared with SoTA baselines across five LLMs on the \textsc{FaithDial} dataset.}
\label{tab3}
\end{table*}

\subsubsection{Automatic Evaluation}
We conducted comprehensive automated evaluation using 9 metrics across 3 dimensions:

\noindent\textbf{Faithfulness.} We employed three metrics: {K-BP} (BERT-Precision between knowledge and response) \cite{chen-etal-2023-fidelity}, {F-Critic} (average entailment score using FaithCritic NLI model) \cite{dziri-etal-2022-faithdial}, and {H-Judge} (faithfulness ratio assessed by HalluJudge LLM) \cite{luo2024halludiallargescalebenchmarkautomatic}.

\noindent\textbf{Expressiveness.} We assessed diversity ({DIV}), context coherence ({COH}), and creative knowledge utilization ({CRE}). {DIV} measures lexical diversity via geometric mean of Distinct-n (n=1,2,3,4) \cite{li-etal-2016-diversity}. {COH} quantifies context-response alignment through cosine similarity of sentence embeddings \cite{su2022contrastive,li-etal-2023-contrastive}. {CRE} evaluates non-extractive knowledge use the \textsc{Coverage} divided by the square root of \textsc{Density} \cite{grusky2020newsroom}. See Appendix \ref{metric} for details.

\noindent\textbf{Quality.} Overall quality was measured using standard overlap-based metrics: {BLEU} \cite{2002-bleu}, {METEOR} \cite{banerjee-lavie-2005-meteor}, and {ROUGE} \cite{lin-2004-rouge}.

\begin{table}[]
\centering
\scalebox{0.9}{
\begin{tabular}{lcccc}
\toprule
Method        & \textbf{Acc}   & \textbf{ROUGE-L} & \textbf{BERT-P} & \textbf{Avg.} \\ \hline
Greedy        & 56.3           & 20.4             & 53.8            & 43.5 \\
Beam          & 58.1           & 21.6             & 55.7            & 45.1 \\
CS            & 55.9           & 19.2             & 52.0            & 42.4 \\
FECS          & 57.6           & \textbf{23.0}    & 57.1            & 45.9 \\ 
F-Nucleus     & 49.5           & 18.8             & 48.9            & 39.1 \\
DoLa          & 56.1           & 20.4             & 53.9            & 43.5 \\
CAD           & 57.4           & \textbf{22.9}    & 56.3            & 45.5 \\ 
\textbf{CoDe} & \textbf{58.8} & 22.4             & \textbf{58.3}   & \textbf{46.5} \\ \bottomrule
\end{tabular}
}
\caption{Evaluation results on the HalluEval (summarization) dataset (Llama2-7B-chat).}
\label{tab:hallueval-summary}
\end{table}

\noindent\textbf{Results.} As shown in Tables \ref{tab1} and \ref{tab5}, CoDe consistently outperforms all ten baseline methods across three faithfulness metrics on all datasets. Our approach also achieves top-2 performance in diversity and relevance metrics. Notably, the CRE scores indicate that CoDe reduces direct knowledge copying compared to other fidelity-enhancing methods like Beam Search and CAD.
We further analyzed knowledge utilization patterns, as shown in Figure \ref{density}. CoDe exhibits lower density than CAD while maintaining higher coverage than sampling methods, indicating substantial token overlap with knowledge sources but minimal contiguous copying. This pattern suggests that CoDe integrates external knowledge more naturally and diversely, extracting relevant information without resorting to verbatim reproduction. Tables 7 and 8 (in Appendix) demonstrate that CoDe performs more closely to the ground-truth in traditional metrics, indicating its overall better performance.
Table \ref{tab3} demonstrates that CoDe significantly improves both fidelity and expressiveness across diverse model architectures and scales. On \textsc{FaithDial}, CoDe achieves H-Judge improvements of +3.9\% for Llama2-7B-chat and +2.4\% for Qwen-2.5-3B-chat over greedy decoding. Remarkably, CoDe enables the 3B model to surpass larger models on multiple metrics (DIV, COH, F-Critic, and H-Judge), highlighting its efficiency in resource-constrained settings.
The results in Tables 3 and 4 demonstrate that CoDe also achieves strong performance on QA and summarization benchmarks that focus solely on faithfulness, highlighting the generalizability of our decoding strategy across diverse task settings.



\begin{table}[]
\centering
\scalebox{0.9}{
\begin{tabular}{lcccc}
\toprule
Method        & \textbf{NQ}    & \textbf{NQ-SWAP} & \textbf{HalluEval(QA)} & \textbf{Avg.} \\ \hline
Greedy        & 32.5           & 26.3             & 54.9                  & 37.9 \\
Beam          & 28.7           & 21.8             & 45.0                  & 31.8 \\
CS            & 30.5           & 22.2             & 52.3                  & 35.0 \\
FECS          & 34.2           & 29.0             & 57.1                  & 40.1 \\ 
F-Nucleus     & 24.4           & 18.7             & 49.6                  & 30.9 \\
DoLa          & 33.5           & 21.4             & 55.8                  & 36.9 \\
CAD           & 34.0           & \textbf{31.9}    & 55.7                  & 40.5 \\ 
\textbf{CoDe} & \textbf{34.5} & 31.6             & \textbf{57.3}         & \textbf{41.1} \\ \bottomrule
\end{tabular}
}
\caption{Accuracy (Acc) results on NQ, NQ-SWAP and HalluEval (QA) datasets (Llama2-7B-chat).}
\label{tab:accuracy}
\end{table}

\begin{figure*}[!h]
  \centerline{\includegraphics[scale=0.45]{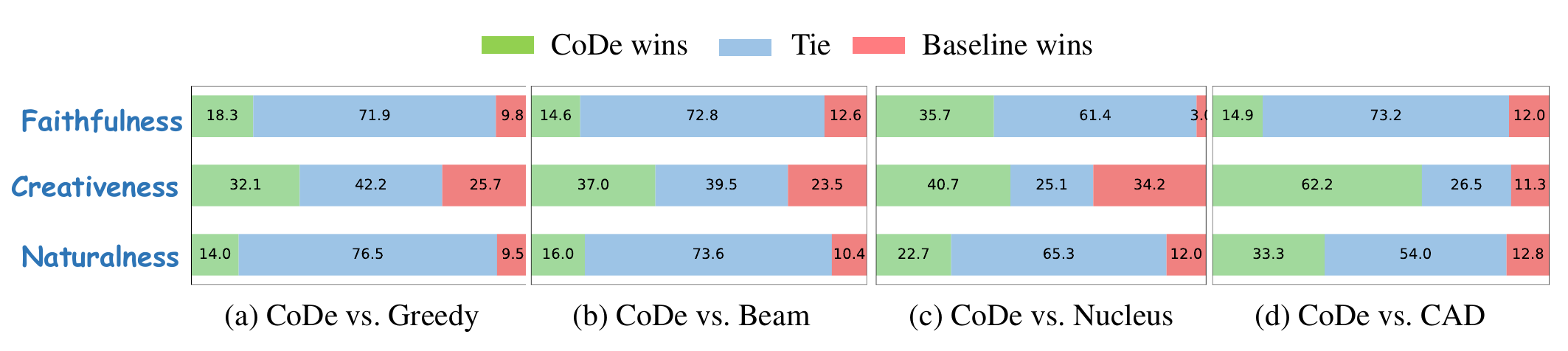}}
  \caption{Human evaluation results on the \textsc{FaithDial} dataset (Llama2-7B-chat). The result is statistically significant with p-value $<$ 0.05, and {Kappa} ($\kappa$) falls between 0.5 and 0.7, suggesting moderate agreement.}
  \label{fighuman}
\end{figure*} 

\begin{figure}[!h]
  \centerline{\includegraphics[scale=0.3]{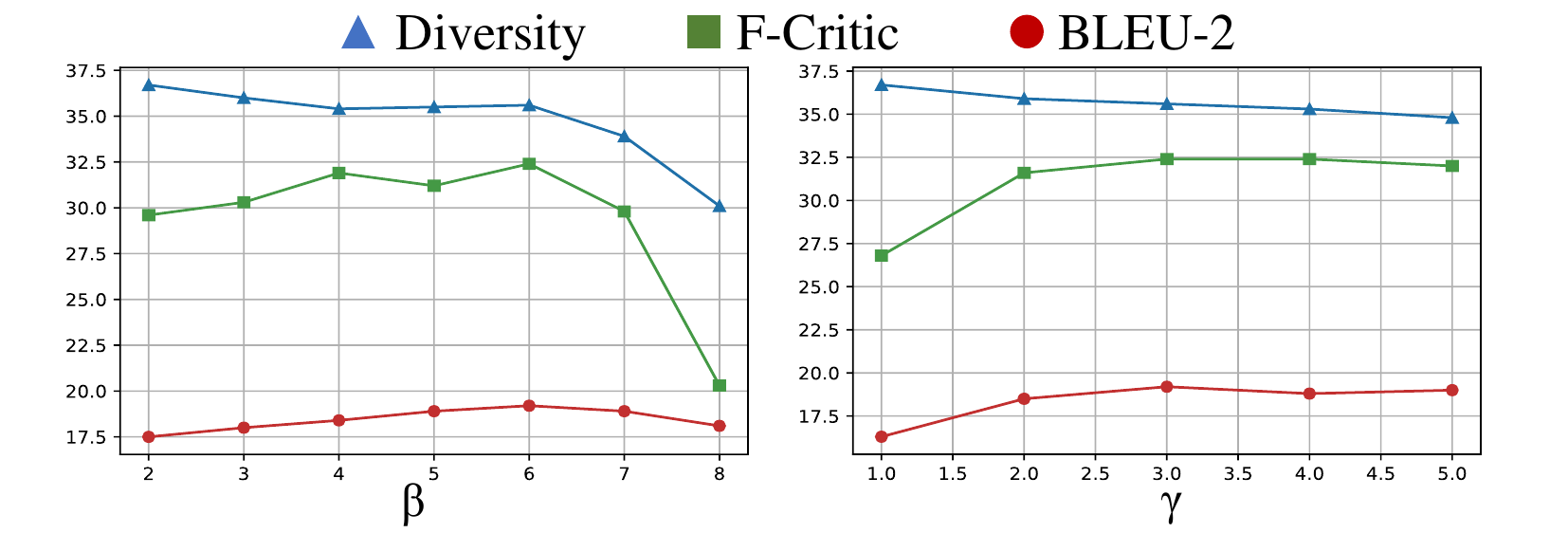}}
  \caption{Hyperparameter study on the \textsc{FaithDial} dataset.}
  \label{ablation-fig}
\end{figure}

\subsubsection{LLMs-based Evaluation}

We employed GPT-4.1 for LLM-as-a-Judge evaluation \cite{G-EVAL,closergeval,zheng2023judgingllmasajudgemtbenchchatbot} on 200 randomly sampled \textsc{FaithDial} test instances. Five decoding methods were evaluated across six criteria (1-5 scale): Naturalness, Coherence, Informativeness, Creativity, Faithfulness, and Factuality, following established rating protocols \cite{GPTscore} (see Appendix \ref{llm-eval}).
Figure \ref{g4-eval} demonstrates that CoDe successfully overcomes the expressiveness-fidelity trade-off, achieving superior overall performance. While nucleus sampling and CAD show bias toward either dimension, CoDe outperforms greedy search across nearly all criteria, confirming the automated evaluation results in Table \ref{tab3}.

\subsubsection{Human Evaluation}

To complement automated and LLM-based evaluations, we conducted human evaluation on 200 randomly selected FaithDial test samples. Five well-educated  annotators compared responses from CoDe and baseline methods across three criteria: Naturalness, Creativity, and Faithfulness (detailed evaluation guidelines are in Figure \ref{fighuman-1}). Inter-annotator agreement was measured using Fleiss' kappa \cite{Fleiss1971MeasuringNS}.
As shown in Figure \ref{fighuman}, CoDe significantly outperformed all baselines in faithfulness. For creativity, annotators preferred CoDe 1.25× over greedy search and 5.5× over CAD. For naturalness, CoDe was favored 1.5× over greedy search and 1.9× over nucleus sampling.

\begin{table}[!t]
\centering
\setlength{\tabcolsep}{4pt} 
\renewcommand{\arraystretch}{1.0} 
\scalebox{0.82}{
\begin{tabular}{lccccccccc}
\toprule
\multicolumn{2}{l}{\multirow{2}{*}{\textbf{Setup}}}& \multicolumn{3}{c}{\textbf{Expressiveness}} & \multicolumn{3}{c}{\textbf{Faithfulness}} & \multirow{2}{*}{\textbf{Avg.}} \\
 \cmidrule(lr){3-5} \cmidrule(lr){6-8}  
 \multicolumn{2}{l}{} & \textbf{DIV} & \textbf{COH} & \textbf{CRE}  & \multicolumn{1}{l}{\textbf{F-Critic}} & \multicolumn{1}{l}{\textbf{H-Judge}} & \multicolumn{1}{l}{\textbf{K-BP}} & \\ \hline
A & CoDe & \underline{35.2} & \underline{57.6} & 29.9 & \textbf{32.4} & \textbf{90.8} & \underline{67.0} & \textbf{52.2} \\
B & -$\alpha$ &  34.9 & 57.5 & \underline{32.1} & {30.1} & {89.2} & {64.7} & 51.4 \\
C& -EOS & 34.7 & 56.8 & 27.3 & {32.3}  & \textbf{90.8} & \textbf{67.3} & \underline{51.5} \\
D&-Sem & 35.0 & 57.1 & 29.6 & 31.4 & 88.6 & 64.1 & 51.0 \\
E&-Att & \underline{35.2} & 57.5 & {30.4} & 30.9 & 88.3 & 63.6 & 51.0 \\
F&-KAR & \textbf{35.6} & \textbf{58.0} & \textbf{33.9} & 29.1 & 85.5 & 59.3 & 50.2 \\ \bottomrule
\end{tabular}
}
\caption{Ablation study on the \textsc{FaithDial} dataset. Avg. denotes the average across all metrics.}
\label{tab:ablation}
\end{table}

\subsection{Ablation Study}
This section presents ablation studies examining key components and hyperparameters ($\beta$, $\gamma$). We evaluated five components on \textsc{FaithDial} using Llama2-7B-chat: Dynamic Fusion Weight (-$\alpha$), Expressiveness-Oriented Stream (-EOS), Knowledge-Aware Reranking (-KAR), semantic reward (-Sem), and attentive reward (-Att).
Table \ref{tab:ablation} shows that all modules are essential. Setup A (equal-weight stream fusion) degrades performance across both dimensions, emphasizing the importance of adaptive internal knowledge utilization. Removing EOS reduces expressiveness similar to other fidelity-focused baselines. Without KAR, expressiveness slightly improves but fidelity drops unacceptably. Setups D and E confirm both reward mechanisms are necessary. Figure \ref{ablation-fig} indicates optimal performance at $\beta$=0.6 and $\gamma$=3. The smooth performance curves across different parameter values demonstrate CoDe's robustness.


\begin{figure}[!t]
  \centerline{\includegraphics[scale=0.33]{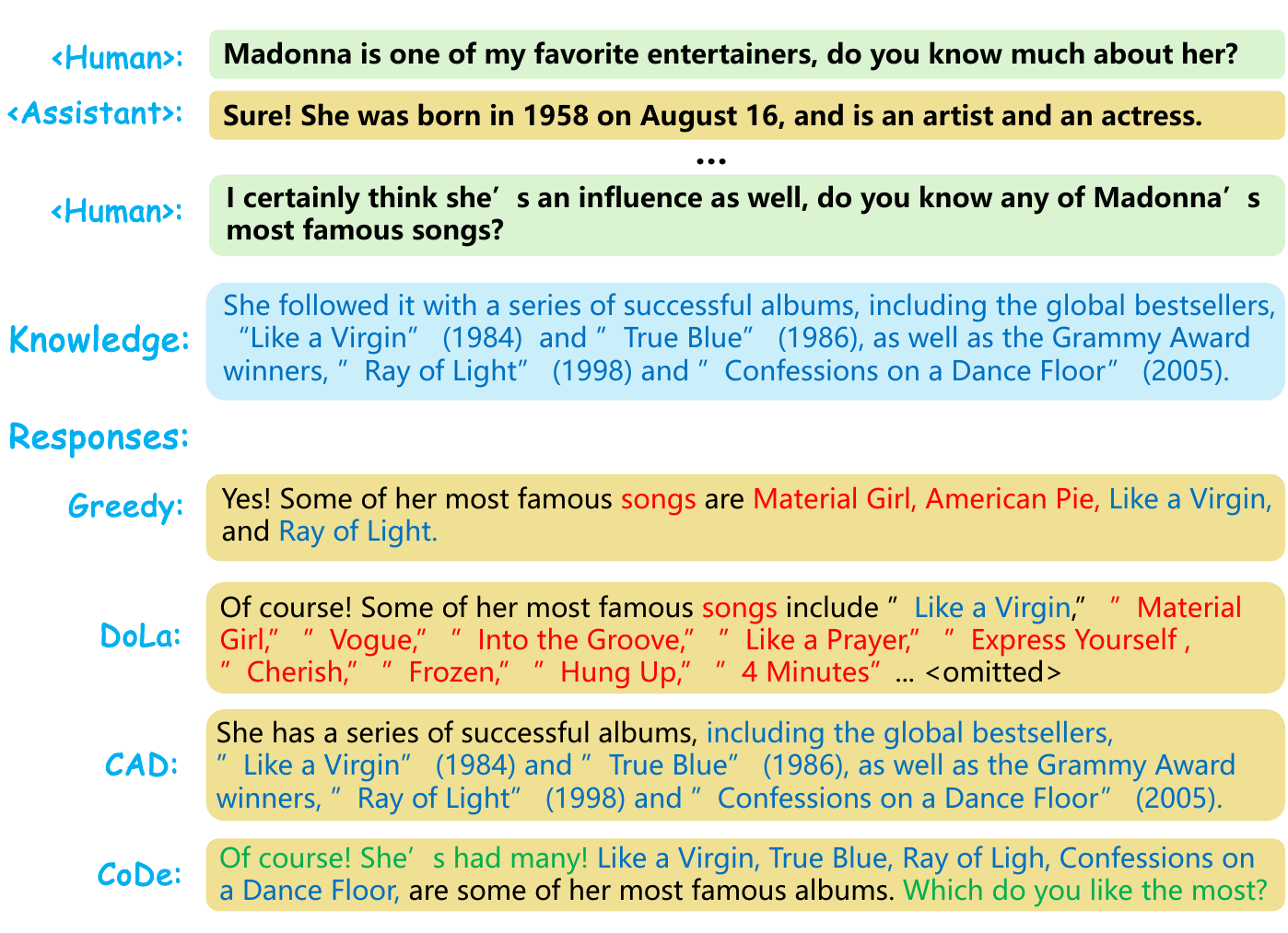}}
  \caption{Case study example with unfaithful hallucinations marked in {\color[HTML]{FE0000}red} and faithful information in {\color[HTML]{3166FF}blue}.}
  \label{case-code}
\end{figure} 

\subsection{Qualitative Examples}
We demonstrate CoDe's ability to generate informative and engaging responses through several case studies. In Figure 8, baseline methods exhibit distinct failure modes: Greedy Search misidentifies Madonna's cover as an original work; both Greedy Search and DoLa confuse song titles with album names, generating unverifiable content; DoLa produces accurate but excessively long lists without proper discourse structure; and CAD merely copies external information without addressing user queries. In contrast, CoDe maintains natural interaction while correctly integrating external knowledge. Additional examples are provided in Appendix Tables 11-13.


\section{Conclusion}

In this paper, we systematically investigated the inherent trade-off between faithfulness and expressiveness in external-knowledge-augmented LLMs. To address this limitation, we introduced Collaborative Decoding (CoDe), a plug-and-play method that dynamically integrates parametric and external knowledge through adaptive dual-stream fusion and knowledge-aware reranking. Extensive experiments across six LLMs and multiple benchmarks demonstrate that CoDe successfully overcomes the faithfulness-expressiveness trade-off. This work opens new avenues for developing decoding strategies that leverage the complementary strengths of internal and external knowledge sources, ultimately advancing the capabilities of LLM assistants in real-world applications.

\bibliography{aaai2026}

\begin{thebibliography}{83}
\providecommand{\natexlab}[1]{#1}

\bibitem[{Asai et~al.(2024)Asai, Wu, Wang, Sil, and Hajishirzi}]{asai2024selfrag}
Asai, A.; Wu, Z.; Wang, Y.; Sil, A.; and Hajishirzi, H. 2024.
\newblock Self-{RAG}: Learning to Retrieve, Generate, and Critique through Self-Reflection.
\newblock In \emph{The Twelfth International Conference on Learning Representations}.

\bibitem[{Bai et~al.(2023)Bai, Bai, Chu, Cui, Dang, Deng, Fan, Ge, Han, Huang, Hui, Ji, Li, Lin, Lin, Liu, Liu, Lu, Lu, Ma, Men, Ren, Ren, Tan, Tan, Tu, Wang, Wang, Wang, Wu, Xu, Xu, Yang, Yang, Yang, Yang, Yao, Yu, Yuan, Yuan, Zhang, Zhang, Zhang, Zhang, Zhou, Zhou, Zhou, and Zhu}]{bai2023qwen}
Bai, J.; Bai, S.; Chu, Y.; Cui, Z.; Dang, K.; Deng, X.; Fan, Y.; Ge, W.; Han, Y.; Huang, F.; Hui, B.; Ji, L.; Li, M.; Lin, J.; Lin, R.; Liu, D.; Liu, G.; Lu, C.; Lu, K.; Ma, J.; Men, R.; Ren, X.; Ren, X.; Tan, C.; Tan, S.; Tu, J.; Wang, P.; Wang, S.; Wang, W.; Wu, S.; Xu, B.; Xu, J.; Yang, A.; Yang, H.; Yang, J.; Yang, S.; Yao, Y.; Yu, B.; Yuan, H.; Yuan, Z.; Zhang, J.; Zhang, X.; Zhang, Y.; Zhang, Z.; Zhou, C.; Zhou, J.; Zhou, X.; and Zhu, T. 2023.
\newblock Qwen Technical Report.
\newblock arXiv:2309.16609.

\bibitem[{Banerjee and Lavie(2005)}]{banerjee-lavie-2005-meteor}
Banerjee, S.; and Lavie, A. 2005.
\newblock {METEOR}: An Automatic Metric for {MT} Evaluation with Improved Correlation with Human Judgments.
\newblock In Goldstein, J.; Lavie, A.; Lin, C.-Y.; and Voss, C., eds., \emph{Proceedings of the {ACL} Workshop on Intrinsic and Extrinsic Evaluation Measures for Machine Translation and/or Summarization}, 65--72. Ann Arbor, Michigan: Association for Computational Linguistics.

\bibitem[{Basu et~al.(2021)Basu, Ramachandran, Keskar, and Varshney}]{Basu2021MirostatAN}
Basu, S.; Ramachandran, G.~S.; Keskar, N.~S.; and Varshney, L.~R. 2021.
\newblock Mirostat: a Neural Text decoding Algorithm that directly controls perplexity.
\newblock In \emph{International Conference on Learning Representations}.

\bibitem[{Boulanger-Lewandowski, Bengio, and Vincent(2013)}]{BoulangerLewandowski2013AudioCR}
Boulanger-Lewandowski, N.; Bengio, Y.; and Vincent, P. 2013.
\newblock Audio Chord Recognition with Recurrent Neural Networks.
\newblock In \emph{International Society for Music Information Retrieval Conference}.

\bibitem[{Chae et~al.(2023)Chae, Song, Ong, Kwon, Kim, Yu, Lee, Kang, and Yeo}]{chae-etal-2023-dialogueDOCTOR}
Chae, H.; Song, Y.; Ong, K.; Kwon, T.; Kim, M.; Yu, Y.; Lee, D.; Kang, D.; and Yeo, J. 2023.
\newblock Dialogue Chain-of-Thought Distillation for Commonsense-aware Conversational Agents.
\newblock In Bouamor, H.; Pino, J.; and Bali, K., eds., \emph{Proceedings of the 2023 Conference on Empirical Methods in Natural Language Processing}, 5606--5632. Singapore: Association for Computational Linguistics.

\bibitem[{Chawla et~al.(2024)Chawla, Rashkin, Tomar, and Reitter}]{chawla-etal-2024-investigating}
Chawla, K.; Rashkin, H.; Tomar, G.~S.; and Reitter, D. 2024.
\newblock Investigating Content Planning for Navigating Trade-offs in Knowledge-Grounded Dialogue.
\newblock In Graham, Y.; and Purver, M., eds., \emph{Proceedings of the 18th Conference of the European Chapter of the Association for Computational Linguistics (Volume 1: Long Papers)}, 2316--2335. St. Julian{'}s, Malta: Association for Computational Linguistics.

\bibitem[{Chen et~al.(2024)Chen, Si, Yang, Liang, Lin, Liu, and Wang}]{chen2024multitaskroleplayingagentcapable}
Chen, S.; Si, Q.; Yang, C.; Liang, Y.; Lin, Z.; Liu, H.; and Wang, W. 2024.
\newblock A Multi-Task Role-Playing Agent Capable of Imitating Character Linguistic Styles.
\newblock arXiv:2411.02457.

\bibitem[{Chen et~al.(2023)Chen, Wu, Chen, and Chen}]{chen-etal-2023-fidelity}
Chen, W.-L.; Wu, C.-K.; Chen, H.-H.; and Chen, C.-C. 2023.
\newblock Fidelity-Enriched Contrastive Search: Reconciling the Faithfulness-Diversity Trade-Off in Text Generation.
\newblock In Bouamor, H.; Pino, J.; and Bali, K., eds., \emph{Proceedings of the 2023 Conference on Empirical Methods in Natural Language Processing}, 843--851. Singapore: Association for Computational Linguistics.

\bibitem[{Chiang and yi~Lee(2023)}]{closergeval}
Chiang, C.-H.; and yi~Lee, H. 2023.
\newblock A Closer Look into Automatic Evaluation Using Large Language Models.
\newblock arXiv:2310.05657.

\bibitem[{Choi et~al.(2023)Choi, Fang, Wang, and Song}]{choi2023kcts}
Choi, S.; Fang, T.; Wang, Z.; and Song, Y. 2023.
\newblock KCTS: Knowledge-Constrained Tree Search Decoding with Token-Level Hallucination Detection.
\newblock arXiv:2310.09044.

\bibitem[{Chuang et~al.(2023)Chuang, Xie, Luo, Kim, Glass, and He}]{chuang2023dola}
Chuang, Y.-S.; Xie, Y.; Luo, H.; Kim, Y.; Glass, J.; and He, P. 2023.
\newblock DoLa: Decoding by Contrasting Layers Improves Factuality in Large Language Models.
\newblock arXiv:2309.03883.

\bibitem[{Dai, Yang, and Si(2025)}]{dai2025sgrpoearlyexitreinforcement}
Dai, M.; Yang, C.; and Si, Q. 2025.
\newblock S-GRPO: Early Exit via Reinforcement Learning in Reasoning Models.
\newblock arXiv:2505.07686.

\bibitem[{Deng et~al.(2023)Deng, Zhang, Huang, and Hu}]{deng-etal-2023-towards-huyue}
Deng, Y.; Zhang, X.; Huang, H.; and Hu, Y. 2023.
\newblock Towards Faithful Dialogues via Focus Learning.
\newblock In Rogers, A.; Boyd-Graber, J.; and Okazaki, N., eds., \emph{Proceedings of the 61st Annual Meeting of the Association for Computational Linguistics (Volume 1: Long Papers)}, 4554--4566. Toronto, Canada: Association for Computational Linguistics.

\bibitem[{Dinan et~al.(2018)Dinan, Roller, Shuster, Fan, Auli, and Weston}]{WoW}
Dinan, E.; Roller, S.; Shuster, K.; Fan, A.; Auli, M.; and Weston, J. 2018.
\newblock Wizard of Wikipedia: Knowledge-Powered Conversational agents.
\newblock \emph{CoRR}, abs/1811.01241.

\bibitem[{Duan et~al.(2023)Duan, Cheng, Wang, Zavalny, Wang, Xu, Kailkhura, and Xu}]{duan2023shifting}
Duan, J.; Cheng, H.; Wang, S.; Zavalny, A.; Wang, C.; Xu, R.; Kailkhura, B.; and Xu, K. 2023.
\newblock Shifting Attention to Relevance: Towards the Uncertainty Estimation of Large Language Models.
\newblock arXiv:2307.01379.

\bibitem[{Dziri et~al.(2022{\natexlab{a}})Dziri, Kamalloo, Milton, Zaiane, Yu, Ponti, and Reddy}]{dziri-etal-2022-faithdial}
Dziri, N.; Kamalloo, E.; Milton, S.; Zaiane, O.; Yu, M.; Ponti, E.~M.; and Reddy, S. 2022{\natexlab{a}}.
\newblock {F}aith{D}ial: A Faithful Benchmark for Information-Seeking Dialogue.
\newblock \emph{Transactions of the Association for Computational Linguistics}, 10: 1473--1490.

\bibitem[{Dziri et~al.(2022{\natexlab{b}})Dziri, Milton, Yu, Zaiane, and Reddy}]{dziri2022origin}
Dziri, N.; Milton, S.; Yu, M.; Zaiane, O.; and Reddy, S. 2022{\natexlab{b}}.
\newblock On the Origin of Hallucinations in Conversational Models: Is it the Datasets or the Models?
\newblock arXiv:2204.07931.

\bibitem[{Fadeeva et~al.(2023)Fadeeva, Vashurin, Tsvigun, Vazhentsev, Petrakov, Fedyanin, Vasilev, Goncharova, Panchenko, Panov, Baldwin, and Shelmanov}]{fadeeva2023lmpolygraph}
Fadeeva, E.; Vashurin, R.; Tsvigun, A.; Vazhentsev, A.; Petrakov, S.; Fedyanin, K.; Vasilev, D.; Goncharova, E.; Panchenko, A.; Panov, M.; Baldwin, T.; and Shelmanov, A. 2023.
\newblock LM-Polygraph: Uncertainty Estimation for Language Models.
\newblock arXiv:2311.07383.

\bibitem[{Fan, Lewis, and Dauphin(2018)}]{fan-etal-2018-topk}
Fan, A.; Lewis, M.; and Dauphin, Y. 2018.
\newblock Hierarchical Neural Story Generation.
\newblock In Gurevych, I.; and Miyao, Y., eds., \emph{Proceedings of the 56th Annual Meeting of the Association for Computational Linguistics (Volume 1: Long Papers)}, 889--898. Melbourne, Australia: Association for Computational Linguistics.

\bibitem[{Fleiss(1971)}]{Fleiss1971MeasuringNS}
Fleiss, J.~L. 1971.
\newblock Measuring nominal scale agreement among many raters.
\newblock \emph{Psychological Bulletin}, 76: 378--382.

\bibitem[{Fu et~al.(2023)Fu, Ng, Jiang, and Liu}]{GPTscore}
Fu, J.; Ng, S.; Jiang, Z.; and Liu, P. 2023.
\newblock GPTScore: Evaluate as You Desire.
\newblock \emph{CoRR}, abs/2302.04166.

\bibitem[{Gao, Yao, and Chen(2021)}]{gao-etal-2021-simcse}
Gao, T.; Yao, X.; and Chen, D. 2021.
\newblock {S}im{CSE}: Simple Contrastive Learning of Sentence Embeddings.
\newblock In Moens, M.-F.; Huang, X.; Specia, L.; and Yih, S. W.-t., eds., \emph{Proceedings of the 2021 Conference on Empirical Methods in Natural Language Processing}, 6894--6910. Online and Punta Cana, Dominican Republic: Association for Computational Linguistics.

\bibitem[{Grattafiori et~al.(2024)Grattafiori, Dubey, Jauhri, Pandey, Kadian, Al-Dahle, Letman, Mathur, Schelten, Vaughan, Yang, Fan, Goyal, Hartshorn, Yang, Mitra, Sravankumar, Korenev, Hinsvark, Rao, Zhang, Rodriguez, Gregerson, Spataru, Roziere, Biron, Tang, Chern, Caucheteux, Nayak, Bi, Marra, McConnell, Keller, Touret, Wu, Wong, Ferrer, Nikolaidis, Allonsius, Song, Pintz, Livshits, Wyatt, Esiobu, Choudhary, Mahajan, Garcia-Olano, Perino, Hupkes, Lakomkin, AlBadawy, Lobanova, Dinan, Smith, Radenovic, Guzmán, Zhang, Synnaeve, Lee, Anderson, Thattai, Nail, Mialon, Pang, Cucurell, Nguyen, Korevaar, Xu, Touvron, Zarov, Ibarra, Kloumann, Misra, Evtimov, Zhang, Copet, Lee, Geffert, Vranes, Park, Mahadeokar, Shah, van~der Linde, Billock, Hong, Lee, Fu, Chi, Huang, Liu, Wang, Yu, Bitton, Spisak, Park, Rocca, Johnstun, Saxe, Jia, Alwala, Prasad, Upasani, Plawiak, Li, Heafield, Stone, El-Arini, Iyer, Malik, Chiu, Bhalla, Lakhotia, Rantala-Yeary, van~der Maaten, Chen, Tan, Jenkins, Martin, Madaan, Malo, Blecher,
  Landzaat, de~Oliveira, Muzzi, Pasupuleti, Singh, Paluri, Kardas, Tsimpoukelli, Oldham, Rita, Pavlova, Kambadur, Lewis, Si, Singh, Hassan, Goyal, Torabi, Bashlykov, Bogoychev, Chatterji, Zhang, Duchenne, Çelebi, Alrassy, Zhang, Li, Vasic, Weng, Bhargava, Dubal, Krishnan, Koura, Xu, He, Dong, Srinivasan, Ganapathy, Calderer, Cabral, Stojnic, Raileanu, Maheswari, Girdhar, Patel, Sauvestre, Polidoro, Sumbaly, Taylor, Silva, Hou, Wang, Hosseini, Chennabasappa, Singh, Bell, Kim, Edunov, Nie, Narang, Raparthy, Shen, Wan, Bhosale, Zhang, Vandenhende, Batra, Whitman, Sootla, Collot, Gururangan, Borodinsky, Herman, Fowler, Sheasha, Georgiou, Scialom, Speckbacher, Mihaylov, Xiao, Karn, Goswami, Gupta, Ramanathan, Kerkez, Gonguet, Do, Vogeti, Albiero, Petrovic, Chu, Xiong, Fu, Meers, Martinet, Wang, Wang, Tan, Xia, Xie, Jia, Wang, Goldschlag, Gaur, Babaei, Wen, Song, Zhang, Li, Mao, Coudert, Yan, Chen, Papakipos, Singh, Srivastava, Jain, Kelsey, Shajnfeld, Gangidi, Victoria, Goldstand, Menon, Sharma, Boesenberg,
  Baevski, Feinstein, Kallet, Sangani, Teo, Yunus, Lupu, Alvarado, Caples, Gu, Ho, Poulton, Ryan, Ramchandani, Dong, Franco, Goyal, Saraf, Chowdhury, Gabriel, Bharambe, Eisenman, Yazdan, James, Maurer, Leonhardi, Huang, Loyd, Paola, Paranjape, Liu, Wu, Ni, Hancock, Wasti, Spence, Stojkovic, Gamido, Montalvo, Parker, Burton, Mejia, Liu, Wang, Kim, Zhou, Hu, Chu, Cai, Tindal, Feichtenhofer, Gao, Civin, Beaty, Kreymer, Li, Adkins, Xu, Testuggine, David, Parikh, Liskovich, Foss, Wang, Le, Holland, Dowling, Jamil, Montgomery, Presani, Hahn, Wood, Le, Brinkman, Arcaute, Dunbar, Smothers, Sun, Kreuk, Tian, Kokkinos, Ozgenel, Caggioni, Kanayet, Seide, Florez, Schwarz, Badeer, Swee, Halpern, Herman, Sizov, Guangyi, Zhang, Lakshminarayanan, Inan, Shojanazeri, Zou, Wang, Zha, Habeeb, Rudolph, Suk, Aspegren, Goldman, Zhan, Damlaj, Molybog, Tufanov, Leontiadis, Veliche, Gat, Weissman, Geboski, Kohli, Lam, Asher, Gaya, Marcus, Tang, Chan, Zhen, Reizenstein, Teboul, Zhong, Jin, Yang, Cummings, Carvill, Shepard, McPhie,
  Torres, Ginsburg, Wang, Wu, U, Saxena, Khandelwal, Zand, Matosich, Veeraraghavan, Michelena, Li, Jagadeesh, Huang, Chawla, Huang, Chen, Garg, A, Silva, Bell, Zhang, Guo, Yu, Moshkovich, Wehrstedt, Khabsa, Avalani, Bhatt, Mankus, Hasson, Lennie, Reso, Groshev, Naumov, Lathi, Keneally, Liu, Seltzer, Valko, Restrepo, Patel, Vyatskov, Samvelyan, Clark, Macey, Wang, Hermoso, Metanat, Rastegari, Bansal, Santhanam, Parks, White, Bawa, Singhal, Egebo, Usunier, Mehta, Laptev, Dong, Cheng, Chernoguz, Hart, Salpekar, Kalinli, Kent, Parekh, Saab, Balaji, Rittner, Bontrager, Roux, Dollar, Zvyagina, Ratanchandani, Yuvraj, Liang, Alao, Rodriguez, Ayub, Murthy, Nayani, Mitra, Parthasarathy, Li, Hogan, Battey, Wang, Howes, Rinott, Mehta, Siby, Bondu, Datta, Chugh, Hunt, Dhillon, Sidorov, Pan, Mahajan, Verma, Yamamoto, Ramaswamy, Lindsay, Lindsay, Feng, Lin, Zha, Patil, Shankar, Zhang, Zhang, Wang, Agarwal, Sajuyigbe, Chintala, Max, Chen, Kehoe, Satterfield, Govindaprasad, Gupta, Deng, Cho, Virk, Subramanian, Choudhury,
  Goldman, Remez, Glaser, Best, Koehler, Robinson, Li, Zhang, Matthews, Chou, Shaked, Vontimitta, Ajayi, Montanez, Mohan, Kumar, Mangla, Ionescu, Poenaru, Mihailescu, Ivanov, Li, Wang, Jiang, Bouaziz, Constable, Tang, Wu, Wang, Wu, Gao, Kleinman, Chen, Hu, Jia, Qi, Li, Zhang, Zhang, Adi, Nam, Yu, Wang, Zhao, Hao, Qian, Li, He, Rait, DeVito, Rosnbrick, Wen, Yang, Zhao, and Ma}]{grattafiori2024llama3herdmodels}
Grattafiori, A.; Dubey, A.; Jauhri, A.; Pandey, A.; Kadian, A.; Al-Dahle, A.; Letman, A.; Mathur, A.; Schelten, A.; Vaughan, A.; Yang, A.; Fan, A.; Goyal, A.; Hartshorn, A.; Yang, A.; Mitra, A.; Sravankumar, A.; Korenev, A.; Hinsvark, A.; Rao, A.; Zhang, A.; Rodriguez, A.; Gregerson, A.; Spataru, A.; Roziere, B.; Biron, B.; Tang, B.; Chern, B.; Caucheteux, C.; Nayak, C.; Bi, C.; Marra, C.; McConnell, C.; Keller, C.; Touret, C.; Wu, C.; Wong, C.; Ferrer, C.~C.; Nikolaidis, C.; Allonsius, D.; Song, D.; Pintz, D.; Livshits, D.; Wyatt, D.; Esiobu, D.; Choudhary, D.; Mahajan, D.; Garcia-Olano, D.; Perino, D.; Hupkes, D.; Lakomkin, E.; AlBadawy, E.; Lobanova, E.; Dinan, E.; Smith, E.~M.; Radenovic, F.; Guzmán, F.; Zhang, F.; Synnaeve, G.; Lee, G.; Anderson, G.~L.; Thattai, G.; Nail, G.; Mialon, G.; Pang, G.; Cucurell, G.; Nguyen, H.; Korevaar, H.; Xu, H.; Touvron, H.; Zarov, I.; Ibarra, I.~A.; Kloumann, I.; Misra, I.; Evtimov, I.; Zhang, J.; Copet, J.; Lee, J.; Geffert, J.; Vranes, J.; Park, J.; Mahadeokar, J.;
  Shah, J.; van~der Linde, J.; Billock, J.; Hong, J.; Lee, J.; Fu, J.; Chi, J.; Huang, J.; Liu, J.; Wang, J.; Yu, J.; Bitton, J.; Spisak, J.; Park, J.; Rocca, J.; Johnstun, J.; Saxe, J.; Jia, J.; Alwala, K.~V.; Prasad, K.; Upasani, K.; Plawiak, K.; Li, K.; Heafield, K.; Stone, K.; El-Arini, K.; Iyer, K.; Malik, K.; Chiu, K.; Bhalla, K.; Lakhotia, K.; Rantala-Yeary, L.; van~der Maaten, L.; Chen, L.; Tan, L.; Jenkins, L.; Martin, L.; Madaan, L.; Malo, L.; Blecher, L.; Landzaat, L.; de~Oliveira, L.; Muzzi, M.; Pasupuleti, M.; Singh, M.; Paluri, M.; Kardas, M.; Tsimpoukelli, M.; Oldham, M.; Rita, M.; Pavlova, M.; Kambadur, M.; Lewis, M.; Si, M.; Singh, M.~K.; Hassan, M.; Goyal, N.; Torabi, N.; Bashlykov, N.; Bogoychev, N.; Chatterji, N.; Zhang, N.; Duchenne, O.; Çelebi, O.; Alrassy, P.; Zhang, P.; Li, P.; Vasic, P.; Weng, P.; Bhargava, P.; Dubal, P.; Krishnan, P.; Koura, P.~S.; Xu, P.; He, Q.; Dong, Q.; Srinivasan, R.; Ganapathy, R.; Calderer, R.; Cabral, R.~S.; Stojnic, R.; Raileanu, R.; Maheswari, R.; Girdhar,
  R.; Patel, R.; Sauvestre, R.; Polidoro, R.; Sumbaly, R.; Taylor, R.; Silva, R.; Hou, R.; Wang, R.; Hosseini, S.; Chennabasappa, S.; Singh, S.; Bell, S.; Kim, S.~S.; Edunov, S.; Nie, S.; Narang, S.; Raparthy, S.; Shen, S.; Wan, S.; Bhosale, S.; Zhang, S.; Vandenhende, S.; Batra, S.; Whitman, S.; Sootla, S.; Collot, S.; Gururangan, S.; Borodinsky, S.; Herman, T.; Fowler, T.; Sheasha, T.; Georgiou, T.; Scialom, T.; Speckbacher, T.; Mihaylov, T.; Xiao, T.; Karn, U.; Goswami, V.; Gupta, V.; Ramanathan, V.; Kerkez, V.; Gonguet, V.; Do, V.; Vogeti, V.; Albiero, V.; Petrovic, V.; Chu, W.; Xiong, W.; Fu, W.; Meers, W.; Martinet, X.; Wang, X.; Wang, X.; Tan, X.~E.; Xia, X.; Xie, X.; Jia, X.; Wang, X.; Goldschlag, Y.; Gaur, Y.; Babaei, Y.; Wen, Y.; Song, Y.; Zhang, Y.; Li, Y.; Mao, Y.; Coudert, Z.~D.; Yan, Z.; Chen, Z.; Papakipos, Z.; Singh, A.; Srivastava, A.; Jain, A.; Kelsey, A.; Shajnfeld, A.; Gangidi, A.; Victoria, A.; Goldstand, A.; Menon, A.; Sharma, A.; Boesenberg, A.; Baevski, A.; Feinstein, A.; Kallet, A.;
  Sangani, A.; Teo, A.; Yunus, A.; Lupu, A.; Alvarado, A.; Caples, A.; Gu, A.; Ho, A.; Poulton, A.; Ryan, A.; Ramchandani, A.; Dong, A.; Franco, A.; Goyal, A.; Saraf, A.; Chowdhury, A.; Gabriel, A.; Bharambe, A.; Eisenman, A.; Yazdan, A.; James, B.; Maurer, B.; Leonhardi, B.; Huang, B.; Loyd, B.; Paola, B.~D.; Paranjape, B.; Liu, B.; Wu, B.; Ni, B.; Hancock, B.; Wasti, B.; Spence, B.; Stojkovic, B.; Gamido, B.; Montalvo, B.; Parker, C.; Burton, C.; Mejia, C.; Liu, C.; Wang, C.; Kim, C.; Zhou, C.; Hu, C.; Chu, C.-H.; Cai, C.; Tindal, C.; Feichtenhofer, C.; Gao, C.; Civin, D.; Beaty, D.; Kreymer, D.; Li, D.; Adkins, D.; Xu, D.; Testuggine, D.; David, D.; Parikh, D.; Liskovich, D.; Foss, D.; Wang, D.; Le, D.; Holland, D.; Dowling, E.; Jamil, E.; Montgomery, E.; Presani, E.; Hahn, E.; Wood, E.; Le, E.-T.; Brinkman, E.; Arcaute, E.; Dunbar, E.; Smothers, E.; Sun, F.; Kreuk, F.; Tian, F.; Kokkinos, F.; Ozgenel, F.; Caggioni, F.; Kanayet, F.; Seide, F.; Florez, G.~M.; Schwarz, G.; Badeer, G.; Swee, G.; Halpern, G.;
  Herman, G.; Sizov, G.; Guangyi; Zhang; Lakshminarayanan, G.; Inan, H.; Shojanazeri, H.; Zou, H.; Wang, H.; Zha, H.; Habeeb, H.; Rudolph, H.; Suk, H.; Aspegren, H.; Goldman, H.; Zhan, H.; Damlaj, I.; Molybog, I.; Tufanov, I.; Leontiadis, I.; Veliche, I.-E.; Gat, I.; Weissman, J.; Geboski, J.; Kohli, J.; Lam, J.; Asher, J.; Gaya, J.-B.; Marcus, J.; Tang, J.; Chan, J.; Zhen, J.; Reizenstein, J.; Teboul, J.; Zhong, J.; Jin, J.; Yang, J.; Cummings, J.; Carvill, J.; Shepard, J.; McPhie, J.; Torres, J.; Ginsburg, J.; Wang, J.; Wu, K.; U, K.~H.; Saxena, K.; Khandelwal, K.; Zand, K.; Matosich, K.; Veeraraghavan, K.; Michelena, K.; Li, K.; Jagadeesh, K.; Huang, K.; Chawla, K.; Huang, K.; Chen, L.; Garg, L.; A, L.; Silva, L.; Bell, L.; Zhang, L.; Guo, L.; Yu, L.; Moshkovich, L.; Wehrstedt, L.; Khabsa, M.; Avalani, M.; Bhatt, M.; Mankus, M.; Hasson, M.; Lennie, M.; Reso, M.; Groshev, M.; Naumov, M.; Lathi, M.; Keneally, M.; Liu, M.; Seltzer, M.~L.; Valko, M.; Restrepo, M.; Patel, M.; Vyatskov, M.; Samvelyan, M.; Clark,
  M.; Macey, M.; Wang, M.; Hermoso, M.~J.; Metanat, M.; Rastegari, M.; Bansal, M.; Santhanam, N.; Parks, N.; White, N.; Bawa, N.; Singhal, N.; Egebo, N.; Usunier, N.; Mehta, N.; Laptev, N.~P.; Dong, N.; Cheng, N.; Chernoguz, O.; Hart, O.; Salpekar, O.; Kalinli, O.; Kent, P.; Parekh, P.; Saab, P.; Balaji, P.; Rittner, P.; Bontrager, P.; Roux, P.; Dollar, P.; Zvyagina, P.; Ratanchandani, P.; Yuvraj, P.; Liang, Q.; Alao, R.; Rodriguez, R.; Ayub, R.; Murthy, R.; Nayani, R.; Mitra, R.; Parthasarathy, R.; Li, R.; Hogan, R.; Battey, R.; Wang, R.; Howes, R.; Rinott, R.; Mehta, S.; Siby, S.; Bondu, S.~J.; Datta, S.; Chugh, S.; Hunt, S.; Dhillon, S.; Sidorov, S.; Pan, S.; Mahajan, S.; Verma, S.; Yamamoto, S.; Ramaswamy, S.; Lindsay, S.; Lindsay, S.; Feng, S.; Lin, S.; Zha, S.~C.; Patil, S.; Shankar, S.; Zhang, S.; Zhang, S.; Wang, S.; Agarwal, S.; Sajuyigbe, S.; Chintala, S.; Max, S.; Chen, S.; Kehoe, S.; Satterfield, S.; Govindaprasad, S.; Gupta, S.; Deng, S.; Cho, S.; Virk, S.; Subramanian, S.; Choudhury, S.;
  Goldman, S.; Remez, T.; Glaser, T.; Best, T.; Koehler, T.; Robinson, T.; Li, T.; Zhang, T.; Matthews, T.; Chou, T.; Shaked, T.; Vontimitta, V.; Ajayi, V.; Montanez, V.; Mohan, V.; Kumar, V.~S.; Mangla, V.; Ionescu, V.; Poenaru, V.; Mihailescu, V.~T.; Ivanov, V.; Li, W.; Wang, W.; Jiang, W.; Bouaziz, W.; Constable, W.; Tang, X.; Wu, X.; Wang, X.; Wu, X.; Gao, X.; Kleinman, Y.; Chen, Y.; Hu, Y.; Jia, Y.; Qi, Y.; Li, Y.; Zhang, Y.; Zhang, Y.; Adi, Y.; Nam, Y.; Yu; Wang; Zhao, Y.; Hao, Y.; Qian, Y.; Li, Y.; He, Y.; Rait, Z.; DeVito, Z.; Rosnbrick, Z.; Wen, Z.; Yang, Z.; Zhao, Z.; and Ma, Z. 2024.
\newblock The Llama 3 Herd of Models.
\newblock arXiv:2407.21783.

\bibitem[{Grusky, Naaman, and Artzi(2020)}]{grusky2020newsroom}
Grusky, M.; Naaman, M.; and Artzi, Y. 2020.
\newblock Newsroom: A Dataset of 1.3 Million Summaries with Diverse Extractive Strategies.
\newblock arXiv:1804.11283.

\bibitem[{Holtzman et~al.(2020)Holtzman, Buys, Du, Forbes, and Choi}]{holtzman2020curious}
Holtzman, A.; Buys, J.; Du, L.; Forbes, M.; and Choi, Y. 2020.
\newblock The Curious Case of Neural Text Degeneration.
\newblock arXiv:1904.09751.

\bibitem[{Huang et~al.(2025)Huang, Yu, Ma, Zhong, Feng, Wang, Chen, Peng, Feng, Qin, and Liu}]{10.1145/3703155survey}
Huang, L.; Yu, W.; Ma, W.; Zhong, W.; Feng, Z.; Wang, H.; Chen, Q.; Peng, W.; Feng, X.; Qin, B.; and Liu, T. 2025.
\newblock A Survey on Hallucination in Large Language Models: Principles, Taxonomy, Challenges, and Open Questions.
\newblock \emph{ACM Trans. Inf. Syst.}, 43(2).

\bibitem[{Huang et~al.(2023)Huang, Song, Wang, Zhao, Chen, Juefei-Xu, and Ma}]{huang2023look}
Huang, Y.; Song, J.; Wang, Z.; Zhao, S.; Chen, H.; Juefei-Xu, F.; and Ma, L. 2023.
\newblock Look Before You Leap: An Exploratory Study of Uncertainty Measurement for Large Language Models.
\newblock arXiv:2307.10236.

\bibitem[{Ji et~al.(2023)Ji, Lee, Frieske, Yu, Su, Xu, Ishii, Bang, Madotto, and Fung}]{Ji_2023}
Ji, Z.; Lee, N.; Frieske, R.; Yu, T.; Su, D.; Xu, Y.; Ishii, E.; Bang, Y.~J.; Madotto, A.; and Fung, P. 2023.
\newblock Survey of Hallucination in Natural Language Generation.
\newblock \emph{ACM Computing Surveys}, 55(12): 1–38.

\bibitem[{Jiang et~al.(2023)Jiang, Sablayrolles, Mensch, Bamford, Chaplot, de~las Casas, Bressand, Lengyel, Lample, Saulnier, Lavaud, Lachaux, Stock, Scao, Lavril, Wang, Lacroix, and Sayed}]{jiang2023mistral7b}
Jiang, A.~Q.; Sablayrolles, A.; Mensch, A.; Bamford, C.; Chaplot, D.~S.; de~las Casas, D.; Bressand, F.; Lengyel, G.; Lample, G.; Saulnier, L.; Lavaud, L.~R.; Lachaux, M.-A.; Stock, P.; Scao, T.~L.; Lavril, T.; Wang, T.; Lacroix, T.; and Sayed, W.~E. 2023.
\newblock Mistral 7B.
\newblock arXiv:2310.06825.

\bibitem[{Kandpal et~al.(2023)Kandpal, Deng, Roberts, Wallace, and Raffel}]{10.5555/3618408.3619049longtai}
Kandpal, N.; Deng, H.; Roberts, A.; Wallace, E.; and Raffel, C. 2023.
\newblock Large language models struggle to learn long-tail knowledge.
\newblock In \emph{Proceedings of the 40th International Conference on Machine Learning}, ICML'23. JMLR.org.

\bibitem[{Kim, Ahn, and Kim(2020)}]{kim2020SKT}
Kim, B.; Ahn, J.; and Kim, G. 2020.
\newblock Sequential Latent Knowledge Selection for Knowledge-Grounded Dialogue.
\newblock arXiv:2002.07510.

\bibitem[{Kwiatkowski et~al.(2019)Kwiatkowski, Palomaki, Redfield, Collins, Parikh, Alberti, Epstein, Polosukhin, Devlin, Lee, Toutanova, Jones, Kelcey, Chang, Dai, Uszkoreit, Le, and Petrov}]{kwiatkowski-etal-2019-natural-NQ}
Kwiatkowski, T.; Palomaki, J.; Redfield, O.; Collins, M.; Parikh, A.; Alberti, C.; Epstein, D.; Polosukhin, I.; Devlin, J.; Lee, K.; Toutanova, K.; Jones, L.; Kelcey, M.; Chang, M.-W.; Dai, A.~M.; Uszkoreit, J.; Le, Q.; and Petrov, S. 2019.
\newblock Natural Questions: A Benchmark for Question Answering Research.
\newblock \emph{Transactions of the Association for Computational Linguistics}, 7: 452--466.

\bibitem[{Lee et~al.(2023)Lee, Ping, Xu, Patwary, Fung, Shoeybi, and Catanzaro}]{lee2023factuality}
Lee, N.; Ping, W.; Xu, P.; Patwary, M.; Fung, P.; Shoeybi, M.; and Catanzaro, B. 2023.
\newblock Factuality Enhanced Language Models for Open-Ended Text Generation.
\newblock arXiv:2206.04624.

\bibitem[{Leng et~al.(2023)Leng, Zhang, Chen, Li, Lu, Miao, and Bing}]{Leng2023MitigatingOHVCD}
Leng, S.; Zhang, H.; Chen, G.; Li, X.; Lu, S.; Miao, C.; and Bing, L. 2023.
\newblock Mitigating Object Hallucinations in Large Vision-Language Models through Visual Contrastive Decoding.
\newblock \emph{2024 IEEE/CVF Conference on Computer Vision and Pattern Recognition (CVPR)}, 13872--13882.

\bibitem[{Li et~al.(2023{\natexlab{a}})Li, Cheng, Zhao, Nie, and Wen}]{li2023halueval}
Li, J.; Cheng, X.; Zhao, W.~X.; Nie, J.-Y.; and Wen, J.-R. 2023{\natexlab{a}}.
\newblock HaluEval: A Large-Scale Hallucination Evaluation Benchmark for Large Language Models.
\newblock arXiv:2305.11747.

\bibitem[{Li et~al.(2016)Li, Galley, Brockett, Gao, and Dolan}]{li-etal-2016-diversity}
Li, J.; Galley, M.; Brockett, C.; Gao, J.; and Dolan, B. 2016.
\newblock A Diversity-Promoting Objective Function for Neural Conversation Models.
\newblock In \emph{Proceedings of the 2016 Conference of the North {A}merican Chapter of the Association for Computational Linguistics: Human Language Technologies}, 110--119. San Diego, California: Association for Computational Linguistics.

\bibitem[{Li et~al.(2023{\natexlab{b}})Li, Patel, Viégas, Pfister, and Wattenberg}]{li2023inferencetime}
Li, K.; Patel, O.; Viégas, F.; Pfister, H.; and Wattenberg, M. 2023{\natexlab{b}}.
\newblock Inference-Time Intervention: Eliciting Truthful Answers from a Language Model.
\newblock arXiv:2306.03341.

\bibitem[{Li et~al.(2023{\natexlab{c}})Li, Holtzman, Fried, Liang, Eisner, Hashimoto, Zettlemoyer, and Lewis}]{li-etal-2023-contrastive}
Li, X.~L.; Holtzman, A.; Fried, D.; Liang, P.; Eisner, J.; Hashimoto, T.; Zettlemoyer, L.; and Lewis, M. 2023{\natexlab{c}}.
\newblock Contrastive Decoding: Open-ended Text Generation as Optimization.
\newblock In Rogers, A.; Boyd-Graber, J.; and Okazaki, N., eds., \emph{Proceedings of the 61st Annual Meeting of the Association for Computational Linguistics (Volume 1: Long Papers)}, 12286--12312. Toronto, Canada: Association for Computational Linguistics.

\bibitem[{Liang et~al.(2024)Liang, Song, Wang, and Zhang}]{liang-etal-2024-learning}
Liang, Y.; Song, Z.; Wang, H.; and Zhang, J. 2024.
\newblock Learning to Trust Your Feelings: Leveraging Self-awareness in {LLM}s for Hallucination Mitigation.
\newblock In Yu, W.; Shi, W.; Yasunaga, M.; Jiang, M.; Zhu, C.; Hajishirzi, H.; Zettlemoyer, L.; and Zhang, Z., eds., \emph{Proceedings of the 3rd Workshop on Knowledge Augmented Methods for NLP}, 44--58. Bangkok, Thailand: Association for Computational Linguistics.

\bibitem[{Lin(2004)}]{lin-2004-rouge}
Lin, C.-Y. 2004.
\newblock {ROUGE}: A Package for Automatic Evaluation of Summaries.
\newblock In \emph{Text Summarization Branches Out}, 74--81. Barcelona, Spain: Association for Computational Linguistics.

\bibitem[{Liu et~al.(2021)Liu, Zhao, Li, Ren, Zhang, and Yin}]{liu2021threestage}
Liu, S.; Zhao, X.; Li, B.; Ren, F.; Zhang, L.; and Yin, S. 2021.
\newblock A Three-Stage Learning Framework for Low-Resource Knowledge-Grounded Dialogue Generation.
\newblock arXiv:2109.04096.

\bibitem[{Liu et~al.(2023)Liu, Iter, Xu, Wang, Xu, and Zhu}]{G-EVAL}
Liu, Y.; Iter, D.; Xu, Y.; Wang, S.; Xu, R.; and Zhu, C. 2023.
\newblock G-Eval: {NLG} Evaluation using {GPT-4} with Better Human Alignment.
\newblock \emph{CoRR}, abs/2303.16634.

\bibitem[{Longpre et~al.(2022)Longpre, Perisetla, Chen, Ramesh, DuBois, and Singh}]{longpre2022entitybasedknowledgeconflictsquestionNQswap}
Longpre, S.; Perisetla, K.; Chen, A.; Ramesh, N.; DuBois, C.; and Singh, S. 2022.
\newblock Entity-Based Knowledge Conflicts in Question Answering.
\newblock arXiv:2109.05052.

\bibitem[{Luo et~al.(2024)Luo, Shen, Li, Peng, Xuan, Wang, and Yang}]{luo2024halludiallargescalebenchmarkautomatic}
Luo, W.; Shen, T.; Li, W.; Peng, G.; Xuan, R.; Wang, H.; and Yang, X. 2024.
\newblock HalluDial: A Large-Scale Benchmark for Automatic Dialogue-Level Hallucination Evaluation.
\newblock arXiv:2406.07070.

\bibitem[{Manakul, Liusie, and Gales(2023)}]{manakul2023selfcheckgpt}
Manakul, P.; Liusie, A.; and Gales, M. J.~F. 2023.
\newblock SelfCheckGPT: Zero-Resource Black-Box Hallucination Detection for Generative Large Language Models.
\newblock arXiv:2303.08896.

\bibitem[{McKenna et~al.(2023)McKenna, Li, Cheng, Hosseini, Johnson, and Steedman}]{mckenna2023sources}
McKenna, N.; Li, T.; Cheng, L.; Hosseini, M.~J.; Johnson, M.; and Steedman, M. 2023.
\newblock Sources of Hallucination by Large Language Models on Inference Tasks.
\newblock arXiv:2305.14552.

\bibitem[{Meng et~al.(2021)Meng, Ren, Chen, Ren, Xi, and Rijke}]{MIKE}
Meng, C.; Ren, P.; Chen, Z.; Ren, Z.; Xi, T.; and Rijke, M.~d. 2021.
\newblock Initiative-Aware Self-Supervised Learning for Knowledge-Grounded Conversations.
\newblock In \emph{Proceedings of the 44th International ACM SIGIR Conference on Research and Development in Information Retrieval}, SIGIR '21, 522–532. New York, NY, USA: Association for Computing Machinery.
\newblock ISBN 9781450380379.

\bibitem[{OpenAI(2023{\natexlab{a}})}]{chatgpt}
OpenAI. 2023{\natexlab{a}}.
\newblock ChatGPT. https://openai.com/blog/chatgpt/.

\bibitem[{OpenAI(2023{\natexlab{b}})}]{openai2023gpt4}
OpenAI. 2023{\natexlab{b}}.
\newblock GPT-4 Technical Report.
\newblock arXiv:2303.08774.

\bibitem[{Pagnoni, Balachandran, and Tsvetkov(2021)}]{pagnoni-etal-2021-understanding-bertscoreP}
Pagnoni, A.; Balachandran, V.; and Tsvetkov, Y. 2021.
\newblock Understanding Factuality in Abstractive Summarization with {FRANK}: A Benchmark for Factuality Metrics.
\newblock In Toutanova, K.; Rumshisky, A.; Zettlemoyer, L.; Hakkani-Tur, D.; Beltagy, I.; Bethard, S.; Cotterell, R.; Chakraborty, T.; and Zhou, Y., eds., \emph{Proceedings of the 2021 Conference of the North American Chapter of the Association for Computational Linguistics: Human Language Technologies}, 4812--4829. Online: Association for Computational Linguistics.

\bibitem[{Papineni et~al.(2002)Papineni, Roukos, Ward, and Zhu}]{2002-bleu}
Papineni, K.; Roukos, S.; Ward, T.; and Zhu, W.-J. 2002.
\newblock {B}leu: a Method for Automatic Evaluation of Machine Translation.
\newblock In \emph{Proceedings of the 40th Annual Meeting of the Association for Computational Linguistics}, 311--318. Philadelphia, Pennsylvania, USA: Association for Computational Linguistics.

\bibitem[{Qwen et~al.(2025)Qwen, :, Yang, Yang, Zhang, Hui, Zheng, Yu, Li, Liu, Huang, Wei, Lin, Yang, Tu, Zhang, Yang, Yang, Zhou, Lin, Dang, Lu, Bao, Yang, Yu, Li, Xue, Zhang, Zhu, Men, Lin, Li, Tang, Xia, Ren, Ren, Fan, Su, Zhang, Wan, Liu, Cui, Zhang, and Qiu}]{qwen2025qwen25technicalreport}
Qwen; :; Yang, A.; Yang, B.; Zhang, B.; Hui, B.; Zheng, B.; Yu, B.; Li, C.; Liu, D.; Huang, F.; Wei, H.; Lin, H.; Yang, J.; Tu, J.; Zhang, J.; Yang, J.; Yang, J.; Zhou, J.; Lin, J.; Dang, K.; Lu, K.; Bao, K.; Yang, K.; Yu, L.; Li, M.; Xue, M.; Zhang, P.; Zhu, Q.; Men, R.; Lin, R.; Li, T.; Tang, T.; Xia, T.; Ren, X.; Ren, X.; Fan, Y.; Su, Y.; Zhang, Y.; Wan, Y.; Liu, Y.; Cui, Z.; Zhang, Z.; and Qiu, Z. 2025.
\newblock Qwen2.5 Technical Report.
\newblock arXiv:2412.15115.

\bibitem[{Ren et~al.(2023)Ren, Wang, Qu, Zhao, Liu, Tian, Wu, rong Wen, and Wang}]{Ren2023InvestigatingTF}
Ren, R.; Wang, Y.; Qu, Y.; Zhao, W.~X.; Liu, J.; Tian, H.; Wu, H.; rong Wen, J.; and Wang, H. 2023.
\newblock Investigating the Factual Knowledge Boundary of Large Language Models with Retrieval Augmentation.
\newblock In \emph{International Conference on Computational Linguistics}.

\bibitem[{Shi et~al.(2024)Shi, Han, Lewis, Tsvetkov, Zettlemoyer, and Yih}]{shi-etal-2024-trustingCAD}
Shi, W.; Han, X.; Lewis, M.; Tsvetkov, Y.; Zettlemoyer, L.; and Yih, W.-t. 2024.
\newblock Trusting Your Evidence: Hallucinate Less with Context-aware Decoding.
\newblock In Duh, K.; Gomez, H.; and Bethard, S., eds., \emph{Proceedings of the 2024 Conference of the North American Chapter of the Association for Computational Linguistics: Human Language Technologies (Volume 2: Short Papers)}, 783--791. Mexico City, Mexico: Association for Computational Linguistics.

\bibitem[{Su et~al.(2022)Su, Lan, Wang, Yogatama, Kong, and Collier}]{su2022contrastive}
Su, Y.; Lan, T.; Wang, Y.; Yogatama, D.; Kong, L.; and Collier, N. 2022.
\newblock A Contrastive Framework for Neural Text Generation.
\newblock arXiv:2202.06417.

\bibitem[{Sun, Ren, and Ren(2023)}]{eacl-2023-generative}
Sun, W.; Ren, P.; and Ren, Z. 2023.
\newblock Generative Knowledge Selection for Knowledge-Grounded Dialogues.
\newblock In \emph{Findings of the Association for Computational Linguistics: EACL 2023}, 2077--2088. Dubrovnik, Croatia: Association for Computational Linguistics.

\bibitem[{Sun et~al.(2023)Sun, Shi, Gao, Ren, de~Rijke, and Ren}]{10.1609/aaai.v37i11.26596sdu}
Sun, W.; Shi, Z.; Gao, S.; Ren, P.; de~Rijke, M.; and Ren, Z. 2023.
\newblock Contrastive learning reduces hallucination in conversations.
\newblock In \emph{Proceedings of the Thirty-Seventh AAAI Conference on Artificial Intelligence and Thirty-Fifth Conference on Innovative Applications of Artificial Intelligence and Thirteenth Symposium on Educational Advances in Artificial Intelligence}, AAAI'23/IAAI'23/EAAI'23. AAAI Press.
\newblock ISBN 978-1-57735-880-0.

\bibitem[{Sutskever, Vinyals, and Le(2014)}]{10.5555/2969033.2969173}
Sutskever, I.; Vinyals, O.; and Le, Q.~V. 2014.
\newblock Sequence to sequence learning with neural networks.
\newblock In \emph{Proceedings of the 28th International Conference on Neural Information Processing Systems - Volume 2}, NIPS'14, 3104–3112. Cambridge, MA, USA: MIT Press.

\bibitem[{Touvron et~al.(2023)Touvron, Martin, Stone, Albert, Almahairi, Babaei, Bashlykov, Batra, Bhargava, Bhosale, Bikel, Blecher, Ferrer, Chen, Cucurull, Esiobu, Fernandes, Fu, Fu, Fuller, Gao, Goswami, Goyal, Hartshorn, Hosseini, Hou, Inan, Kardas, Kerkez, Khabsa, Kloumann, Korenev, Koura, Lachaux, Lavril, Lee, Liskovich, Lu, Mao, Martinet, Mihaylov, Mishra, Molybog, Nie, Poulton, Reizenstein, Rungta, Saladi, Schelten, Silva, Smith, Subramanian, Tan, Tang, Taylor, Williams, Kuan, Xu, Yan, Zarov, Zhang, Fan, Kambadur, Narang, Rodriguez, Stojnic, Edunov, and Scialom}]{touvron2023llama2}
Touvron, H.; Martin, L.; Stone, K.; Albert, P.; Almahairi, A.; Babaei, Y.; Bashlykov, N.; Batra, S.; Bhargava, P.; Bhosale, S.; Bikel, D.; Blecher, L.; Ferrer, C.~C.; Chen, M.; Cucurull, G.; Esiobu, D.; Fernandes, J.; Fu, J.; Fu, W.; Fuller, B.; Gao, C.; Goswami, V.; Goyal, N.; Hartshorn, A.; Hosseini, S.; Hou, R.; Inan, H.; Kardas, M.; Kerkez, V.; Khabsa, M.; Kloumann, I.; Korenev, A.; Koura, P.~S.; Lachaux, M.-A.; Lavril, T.; Lee, J.; Liskovich, D.; Lu, Y.; Mao, Y.; Martinet, X.; Mihaylov, T.; Mishra, P.; Molybog, I.; Nie, Y.; Poulton, A.; Reizenstein, J.; Rungta, R.; Saladi, K.; Schelten, A.; Silva, R.; Smith, E.~M.; Subramanian, R.; Tan, X.~E.; Tang, B.; Taylor, R.; Williams, A.; Kuan, J.~X.; Xu, P.; Yan, Z.; Zarov, I.; Zhang, Y.; Fan, A.; Kambadur, M.; Narang, S.; Rodriguez, A.; Stojnic, R.; Edunov, S.; and Scialom, T. 2023.
\newblock Llama 2: Open Foundation and Fine-Tuned Chat Models.
\newblock arXiv:2307.09288.

\bibitem[{Wang et~al.(2022)Wang, Li, Lin, Meng, Yang, Wang, and Zhou}]{Wang2022seek}
Wang, L.; Li, J.; Lin, Z.; Meng, F.; Yang, C.; Wang, W.; and Zhou, J. 2022.
\newblock Empathetic Dialogue Generation via Sensitive Emotion Recognition and Sensible Knowledge Selection.
\newblock In \emph{Conference on Empirical Methods in Natural Language Processing}.

\bibitem[{Wang et~al.(2025)Wang, Li, Yang, Lin, Tang, Liu, Cao, Wang, and Wang}]{wang-etal-2025-sibyl}
Wang, L.; Li, J.; Yang, C.; Lin, Z.; Tang, H.; Liu, H.; Cao, Y.; Wang, J.; and Wang, W. 2025.
\newblock Sibyl: Empowering Empathetic Dialogue Generation in Large Language Models via Sensible and Visionary Commonsense Inference.
\newblock In Rambow, O.; Wanner, L.; Apidianaki, M.; Al-Khalifa, H.; Eugenio, B.~D.; and Schockaert, S., eds., \emph{Proceedings of the 31st International Conference on Computational Linguistics}, 123--140. Abu Dhabi, UAE: Association for Computational Linguistics.

\bibitem[{Wei et~al.(2022)Wei, Wang, Schuurmans, Bosma, Ichter, Xia, Chi, Le, and Zhou}]{10.5555/3600270.3602070COT}
Wei, J.; Wang, X.; Schuurmans, D.; Bosma, M.; Ichter, B.; Xia, F.; Chi, E.~H.; Le, Q.~V.; and Zhou, D. 2022.
\newblock Chain-of-thought prompting elicits reasoning in large language models.
\newblock In \emph{Proceedings of the 36th International Conference on Neural Information Processing Systems}, NIPS '22. Red Hook, NY, USA: Curran Associates Inc.
\newblock ISBN 9781713871088.

\bibitem[{Wolf et~al.(2020)Wolf, Debut, Sanh, Chaumond, Delangue, Moi, Cistac, Rault, Louf, Funtowicz, Davison, Shleifer, von Platen, Ma, Jernite, Plu, Xu, Le~Scao, Gugger, Drame, Lhoest, and Rush}]{transformers}
Wolf, T.; Debut, L.; Sanh, V.; Chaumond, J.; Delangue, C.; Moi, A.; Cistac, P.; Rault, T.; Louf, R.; Funtowicz, M.; Davison, J.; Shleifer, S.; von Platen, P.; Ma, C.; Jernite, Y.; Plu, J.; Xu, C.; Le~Scao, T.; Gugger, S.; Drame, M.; Lhoest, Q.; and Rush, A. 2020.
\newblock Transformers: State-of-the-Art Natural Language Processing.
\newblock In \emph{Proceedings of the 2020 Conference on Empirical Methods in Natural Language Processing: System Demonstrations}, 38--45. Online: Association for Computational Linguistics.

\bibitem[{Xia et~al.(2023)Xia, Gao, Zeng, and Chen}]{xia2023sheared}
Xia, M.; Gao, T.; Zeng, Z.; and Chen, D. 2023.
\newblock Sheared llama: Accelerating language model pre-training via structured pruning.
\newblock \emph{arXiv preprint arXiv:2310.06694}.

\bibitem[{Xu et~al.(2022)Xu, Zhou, Fu, Kan, and Ng}]{xu-etal-2022-corefdiffs}
Xu, L.; Zhou, Q.; Fu, J.; Kan, M.-Y.; and Ng, S.-K. 2022.
\newblock {C}oref{D}iffs: Co-referential and Differential Knowledge Flow in Document Grounded Conversations.
\newblock In \emph{Proceedings of the 29th International Conference on Computational Linguistics}, 471--484. Gyeongju, Republic of Korea: International Committee on Computational Linguistics.

\bibitem[{Yang et~al.(2023{\natexlab{a}})Yang, Xiao, Wang, Zhang, Bian, Yin, Lv, Pan, Wang, Yan, Yang, Deng, Wang, Liu, Ai, Dong, Zhao, Xu, Sun, Zhang, Liu, Ji, Xie, Dai, Fang, Su, Song, Liu, Ru, Ma, Wang, Liu, Lin, Nie, Guo, Sun, Zhang, Li, Li, Cheng, Chen, Zeng, Wang, Chen, Men, Yu, Pan, Shen, Wang, Li, Jiang, Gao, Zhang, Zhou, and Wu}]{yang2023baichuan}
Yang, A.; Xiao, B.; Wang, B.; Zhang, B.; Bian, C.; Yin, C.; Lv, C.; Pan, D.; Wang, D.; Yan, D.; Yang, F.; Deng, F.; Wang, F.; Liu, F.; Ai, G.; Dong, G.; Zhao, H.; Xu, H.; Sun, H.; Zhang, H.; Liu, H.; Ji, J.; Xie, J.; Dai, J.; Fang, K.; Su, L.; Song, L.; Liu, L.; Ru, L.; Ma, L.; Wang, M.; Liu, M.; Lin, M.; Nie, N.; Guo, P.; Sun, R.; Zhang, T.; Li, T.; Li, T.; Cheng, W.; Chen, W.; Zeng, X.; Wang, X.; Chen, X.; Men, X.; Yu, X.; Pan, X.; Shen, Y.; Wang, Y.; Li, Y.; Jiang, Y.; Gao, Y.; Zhang, Y.; Zhou, Z.; and Wu, Z. 2023{\natexlab{a}}.
\newblock Baichuan 2: Open Large-scale Language Models.
\newblock arXiv:2309.10305.

\bibitem[{Yang et~al.(2024)Yang, Jia, Gu, Lin, Chen, Pang, Yin, Sun, Wu, and Wang}]{yang2024orthogonalfinetuningdirectpreference}
Yang, C.; Jia, R.; Gu, N.; Lin, Z.; Chen, S.; Pang, C.; Yin, W.; Sun, Y.; Wu, H.; and Wang, W. 2024.
\newblock Orthogonal Finetuning for Direct Preference Optimization.
\newblock arXiv:2409.14836.

\bibitem[{Yang et~al.(2022)Yang, Lin, Li, Meng, Wang, Wang, and Zhou}]{yang-etal-2022-take}
Yang, C.; Lin, Z.; Li, J.; Meng, F.; Wang, W.; Wang, L.; and Zhou, J. 2022.
\newblock {TAKE}: Topic-shift Aware Knowledge s{E}lection for Dialogue Generation.
\newblock In \emph{Proceedings of the 29th International Conference on Computational Linguistics}, 253--265. Gyeongju, Republic of Korea: International Committee on Computational Linguistics.

\bibitem[{Yang et~al.(2023{\natexlab{b}})Yang, Lin, Wang, Tian, Pang, Li, Ho, Cao, and Wang}]{yang-etal-2023-multi-level}
Yang, C.; Lin, Z.; Wang, L.; Tian, C.; Pang, L.; Li, J.; Ho, Q.; Cao, Y.; and Wang, W. 2023{\natexlab{b}}.
\newblock Multi-level Adaptive Contrastive Learning for Knowledge Internalization in Dialogue Generation.
\newblock arXiv:2310.08943.

\bibitem[{Yang et~al.(2025{\natexlab{a}})Yang, Si, Dai, Yao, Zheng, Chen, Lin, and Wang}]{yang2025testtimepromptintervention}
Yang, C.; Si, Q.; Dai, M.; Yao, D.; Zheng, M.; Chen, M.; Lin, Z.; and Wang, W. 2025{\natexlab{a}}.
\newblock Test-time Prompt Intervention.
\newblock arXiv:2508.02511.

\bibitem[{Yang et~al.(2025{\natexlab{b}})Yang, Si, Duan, Zhu, Zhu, Li, Lin, Cao, and Wang}]{yang2025dynamicearlyexitreasoning}
Yang, C.; Si, Q.; Duan, Y.; Zhu, Z.; Zhu, C.; Li, Q.; Lin, Z.; Cao, L.; and Wang, W. 2025{\natexlab{b}}.
\newblock Dynamic Early Exit in Reasoning Models.
\newblock arXiv:2504.15895.

\bibitem[{Yu et~al.(2024)Yu, Wu, Chen, and Zhou}]{yu-etal-2024-llms}
Yu, J.; Wu, S.; Chen, J.; and Zhou, W. 2024.
\newblock {LLM}s as Collaborator: Demands-Guided Collaborative Retrieval-Augmented Generation for Commonsense Knowledge-Grounded Open-Domain Dialogue Systems.
\newblock In Al-Onaizan, Y.; Bansal, M.; and Chen, Y.-N., eds., \emph{Findings of the Association for Computational Linguistics: EMNLP 2024}, 13586--13612. Miami, Florida, USA: Association for Computational Linguistics.

\bibitem[{Zhan et~al.(2021)Zhan, Shen, Chen, and Zhang}]{2021-colv}
Zhan, H.; Shen, L.; Chen, H.; and Zhang, H. 2021.
\newblock {C}o{LV}: A Collaborative Latent Variable Model for Knowledge-Grounded Dialogue Generation.
\newblock In \emph{Proceedings of the 2021 Conference on Empirical Methods in Natural Language Processing}, 2250--2261. Online and Punta Cana, Dominican Republic: Association for Computational Linguistics.

\bibitem[{Zhang et~al.(2023{\natexlab{a}})Zhang, Qiu, Guo, Deng, Zhang, Zhang, Zhou, Wang, and Fu}]{zhang2023enhancing}
Zhang, T.; Qiu, L.; Guo, Q.; Deng, C.; Zhang, Y.; Zhang, Z.; Zhou, C.; Wang, X.; and Fu, L. 2023{\natexlab{a}}.
\newblock Enhancing Uncertainty-Based Hallucination Detection with Stronger Focus.
\newblock arXiv:2311.13230.

\bibitem[{Zhang et~al.(2023{\natexlab{b}})Zhang, Qiu, Guo, Deng, Zhang, Zhang, Zhou, Wang, and Fu}]{zhang-etal-2023-enhancing-uncertainty}
Zhang, T.; Qiu, L.; Guo, Q.; Deng, C.; Zhang, Y.; Zhang, Z.; Zhou, C.; Wang, X.; and Fu, L. 2023{\natexlab{b}}.
\newblock Enhancing Uncertainty-Based Hallucination Detection with Stronger Focus.
\newblock In Bouamor, H.; Pino, J.; and Bali, K., eds., \emph{Proceedings of the 2023 Conference on Empirical Methods in Natural Language Processing}, 915--932. Singapore: Association for Computational Linguistics.

\bibitem[{Zhang et~al.(2024)Zhang, Peng, Tian, Zhou, Jin, Song, Mi, and Meng}]{zhang-etal-2024-selfalign}
Zhang, X.; Peng, B.; Tian, Y.; Zhou, J.; Jin, L.; Song, L.; Mi, H.; and Meng, H. 2024.
\newblock Self-Alignment for Factuality: Mitigating Hallucinations in {LLM}s via Self-Evaluation.
\newblock In Ku, L.-W.; Martins, A.; and Srikumar, V., eds., \emph{Proceedings of the 62nd Annual Meeting of the Association for Computational Linguistics (Volume 1: Long Papers)}, 1946--1965. Bangkok, Thailand: Association for Computational Linguistics.

\bibitem[{Zhang et~al.(2023{\natexlab{c}})Zhang, Li, Cui, Cai, Liu, Fu, Huang, Zhao, Zhang, Chen, Wang, Luu, Bi, Shi, and Shi}]{zhang2023sirens}
Zhang, Y.; Li, Y.; Cui, L.; Cai, D.; Liu, L.; Fu, T.; Huang, X.; Zhao, E.; Zhang, Y.; Chen, Y.; Wang, L.; Luu, A.~T.; Bi, W.; Shi, F.; and Shi, S. 2023{\natexlab{c}}.
\newblock Siren's Song in the AI Ocean: A Survey on Hallucination in Large Language Models.
\newblock arXiv:2309.01219.

\bibitem[{Zhao et~al.(2020{\natexlab{a}})Zhao, Wu, Tao, Xu, Zhao, and Yan}]{zhao2020lowresource}
Zhao, X.; Wu, W.; Tao, C.; Xu, C.; Zhao, D.; and Yan, R. 2020{\natexlab{a}}.
\newblock Low-Resource Knowledge-Grounded Dialogue Generation.
\newblock arXiv:2002.10348.

\bibitem[{Zhao et~al.(2020{\natexlab{b}})Zhao, Wu, Xu, Tao, Zhao, and Yan}]{zhao2020knowledgegrounded}
Zhao, X.; Wu, W.; Xu, C.; Tao, C.; Zhao, D.; and Yan, R. 2020{\natexlab{b}}.
\newblock Knowledge-Grounded Dialogue Generation with Pre-trained Language Models.
\newblock arXiv:2010.08824.

\bibitem[{Zheng et~al.(2023)Zheng, Chiang, Sheng, Zhuang, Wu, Zhuang, Lin, Li, Li, Xing, Zhang, Gonzalez, and Stoica}]{zheng2023judgingllmasajudgemtbenchchatbot}
Zheng, L.; Chiang, W.-L.; Sheng, Y.; Zhuang, S.; Wu, Z.; Zhuang, Y.; Lin, Z.; Li, Z.; Li, D.; Xing, E.~P.; Zhang, H.; Gonzalez, J.~E.; and Stoica, I. 2023.
\newblock Judging LLM-as-a-Judge with MT-Bench and Chatbot Arena.
\newblock arXiv:2306.05685.

\bibitem[{Zheng, Milic-Frayling, and Zhou(2021)}]{zheng-etal-2021-knowledge}
Zheng, W.; Milic-Frayling, N.; and Zhou, K. 2021.
\newblock Knowledge-Grounded Dialogue Generation with Term-level De-noising.
\newblock In Zong, C.; Xia, F.; Li, W.; and Navigli, R., eds., \emph{Findings of the Association for Computational Linguistics: ACL-IJCNLP 2021}, 2972--2983. Online: Association for Computational Linguistics.

\bibitem[{Zhou et~al.(2022)Zhou, Gopalakrishnan, Hedayatnia, Kim, Pujara, Ren, Liu, and Hakkani-Tur}]{zhou-etal-2022-think}
Zhou, P.; Gopalakrishnan, K.; Hedayatnia, B.; Kim, S.; Pujara, J.; Ren, X.; Liu, Y.; and Hakkani-Tur, D. 2022.
\newblock Think Before You Speak: Explicitly Generating Implicit Commonsense Knowledge for Response Generation.
\newblock In Muresan, S.; Nakov, P.; and Villavicencio, A., eds., \emph{Proceedings of the 60th Annual Meeting of the Association for Computational Linguistics (Volume 1: Long Papers)}, 1237--1252. Dublin, Ireland: Association for Computational Linguistics.

\end{thebibliography}

\appendix

\begin{table*}[]
\centering
\setlength{\tabcolsep}{5pt} 
\renewcommand{\arraystretch}{0.95} 
\scalebox{0.86}{
\begin{tabular}{lccccccccccccccc}
\toprule
  \multirow{3}{*}{Method} & \multicolumn{7}{c}{\textbf{WoW-Seen}} & \multicolumn{7}{c}{\textbf{WoW-Unseen}} \\ 
 \cmidrule(lr){2-8} \cmidrule(lr){9-15} 
 & \multicolumn{3}{c}{\textbf{Expressiveness}} & \multicolumn{3}{c}{\textbf{Faithfulness}} & \multirow{2}{*}{\textbf{Avg.}} & \multicolumn{3}{c}{\textbf{Expressiveness}} & \multicolumn{3}{c}{\textbf{Faithfulness}} & \multirow{2}{*}{\textbf{Avg.}} \\ 
 \cmidrule(lr){2-4}  \cmidrule(lr){5-7} \cmidrule(lr){9-11}  \cmidrule(lr){12-14}
 & \textbf{DIV} & \textbf{COH} & \textbf{CRE} & \multicolumn{1}{l}{\textbf{F-Critic}} & \multicolumn{1}{l}{\textbf{H-Judge}} & \multicolumn{1}{l}{\textbf{K-BP}} & & \textbf{DIV} & \textbf{COH} & \textbf{CRE} & \multicolumn{1}{l}{\textbf{F-Critic}} & \multicolumn{1}{l}{\textbf{H-Judge}} & \multicolumn{1}{l}{\textbf{K-BP}} & \\ \hline
Greedy & 31.5 & {58.3} & 28.7 & 19.5 & 84.2 & 58.1 & 46.7
& 22.4 & {58.2} & 28.8 & 17.6 & 86.7 & 58.6 & 45.4 \\
Beam & 30.7 & {58.6} & 24.1 & 22.1 & 86.1 & {62.4} & 47.3
& 21.4 & \textbf{58.9} & 24.0 & \underline{22.9} & 87.6 & 62.8 & 46.3 \\
CS & 32.0 & 58.4 & 29.0 & 21.0 & 84.2 & 57.6 & 47.0
& 23.0 & {57.5} & 28.6 & 18.9 & 85.9 & 58.4 & 45.4 \\
FECS & 33.4 & 57.8 & 28.1 & 23.5 & {86.3} & 61.9 & \underline{48.5}
& 23.3 & 57.2 & 28.6 & \underline{22.9} & \underline{88.6} & 61.9 & \underline{47.1} \\ 
top-k & \textbf{35.8} & 58.2 & \textbf{33.4} & 15.0 & 66.5 & 53.9 & 43.8
& \underline{29.8} & 57.7 & \textbf{33.6} & 14.2 & 70.1 & 54.5 & 43.3 \\
Nucleus & {34.9} & 58.4 & \underline{33.1} & 16.3 & 73.6 & 54.9 & 45.2
& 28.2 & 57.8 & {32.8} & 15.9 & 76.8 & 55.5 & 44.5 \\
F-Nucleus & 33.9 & \underline{58.7} & 31.3 & 16.7 & 77.6 & 56.1 & 45.7
& 26.6 & 58.4 & 31.3 & 17.5 & 80.0 & 56.6 & 45.1 \\ 
CD & 34.7 & 57.1 & 32.9 & 16.3 & 68.9 & 57.0 & 44.5
& 34.9 & 56.9 & \underline{33.0} & 13.7 & 73.8 & 57.5 & 45.0 \\
DoLa & 33.1 & 58.0 & 30.4 & 22.9 & 84.2 & 57.9 & 47.8
& 23.9 & 57.5 & 31.3 & 21.6 & 85.9 & 58.4 & 46.4 \\
CAD & 28.1 & 53.9 & 18.6 & \underline{26.4} & \underline{86.7} & \textbf{65.6} & 46.5
& 20.7 & 53.6 & 21.8 & {22.6} & 87.7 & \underline{64.8} & 45.2 \\ 
\textbf{CoDe} & \underline{35.2} & \textbf{58.9} & 27.7 & \textbf{26.8} & \textbf{88.0} & \underline{65.0} & \textbf{50.3}
& \textbf{30.1} & \underline{58.6} & 28.1 & \textbf{25.6} & \textbf{89.7} & \textbf{65.2} & \textbf{49.6} \\ 
\bottomrule
\end{tabular}
}
\caption{Automatic evaluation results on the WoW dataset (Llama2-7B-chat). Avg. denotes the average across all metrics.}
\label{tab5}
\end{table*}

\begin{table}[]
\centering
\scalebox{0.86}{
\begin{tabular}{lcccc}
\toprule
Method        & \textbf{BLEU-2/4}    & \textbf{METEOR} & \textbf{ROUGE-L} & \textbf{Avg.} \\ \hline
Greedy        & 16.5/7.8            & 20.2            & 27.6            & 18.0 \\
Beam          & 16.6/8.3            & 22.1            & 28.6            & 18.9 \\
CS            & 16.4/7.8            & 20.0            & 27.5            & 17.9 \\
FECS          & 18.3/8.8            & 20.9            & 28.7            & 19.2 \\ 
top-k         & 13.4/5.7            & 17.7            & 23.7            & 15.1 \\
Nucleus       & 14.2/6.3            & 18.5            & 24.6            & 15.9 \\
F-Nucleus     & 14.9/6.5            & 19.2            & 25.4            & 16.5 \\ 
CD            & 12.4/5.6            & 16.8            & 23.9            & 14.7 \\
DoLa          & 17.1/8.1            & 19.6            & 27.4            & 18.1 \\
CAD           & 18.5/8.9            & 22.2            & 28.0            & 19.4 \\ 
\textbf{CoDe} & \textbf{18.5/9.1} & \textbf{22.5}  & \textbf{28.9}   & \textbf{19.8} \\ \bottomrule
\end{tabular}
}
\caption{Automatic Evaluation results on the HalluDial dataset (Llama2-7B-chat). Avg. denotes the average across all metrics.}
\label{tab6}
\end{table}

\begin{table}[]
\centering
\setlength{\tabcolsep}{5pt}
\renewcommand{\arraystretch}{0.95}
\scalebox{0.9}{
\begin{tabular}{lcccc}
\toprule
Method        & \textbf{BLEU-2/4}    & \textbf{METEOR} & \textbf{ROUGE-L} & \textbf{Avg.} \\ \hline
Greedy        & 17.0/8.0            & 20.1            & 27.1            & 18.1 \\
Beam          & 17.7/8.7            & 21.0            & 28.2            & 18.9 \\
CS            & 16.4/7.6            & 18.7            & 25.7            & 17.1 \\
FECS          & 18.4/8.8            & 20.8            & 28.1            & 19.0 \\ 
top-k         & 14.3/6.2            & 18.2            & 24.0            & 15.7 \\
Nucleus       & 14.8/6.4            & 18.5            & 24.5            & 16.1 \\
F-Nucleus     & 15.6/7.0            & 18.9            & 25.4            & 16.7 \\ 
CD            & 13.6/5.9            & 17.6            & 24.5            & 15.4 \\
DoLa          & 18.1/8.7            & 20.0            & 27.7            & 18.6 \\
CAD           & 18.6/8.9            & 21.2            & 27.6            & 19.1 \\ 
\textbf{CoDe} & \textbf{19.2/9.5} & \textbf{22.6}  & \textbf{29.3}   & \textbf{20.1} \\ \bottomrule
\end{tabular}
}
\caption{Automatic Evaluation results on the \textsc{FaithDial} dataset (Llama2-7B-chat).}
\label{tab2}
\end{table}

\section{Preliminary Experimental Setups.}

In the example shown in Table \ref{tab:case1}, Greedy Search mistakenly treated Madonna's cover song as her original work, leading to a factual mistake. Greedy Search, Nucleus Sampling, and DoLa all confused song titles with album names, and generated many songs that cannot be verified by external knowledge. Although DoLa indeed activated the model's internal knowledge to produce numerous accurate song titles, it kept listing them endlessly, resulting in an excessively long response. CAD failed to respond to the user's query and simply copied external information into the reply, showing weak expression capabilities. By comparison, CoDe not only interacted with the user, but also correctly utilized external knowledge.

\section{Datasets} \label{dataset}

\textbf{WoW} is collected based on Wikipedia, with one crowd-sourcer acts as a knowledgeable wizard and the other plays the role of an inquisitive apprentice. The objective is to generate responses based on given knowledge snippets, taken from Wikipedia, that are pertinent to the conversation topic. The ground-truth responses in the dataset are annotated by humans based on the best knowledge they selected. We evaluated all the decoding methods on both the test {seen} and unseen set. The test {seen} set includes 4,336 samples where the topics were seen in the training set, while the unseen test set includes 4,370 samples where the topics were not seen in the training set \cite{WoW}.

\begin{figure}[!t]
  \centerline{\includegraphics[scale=0.23]{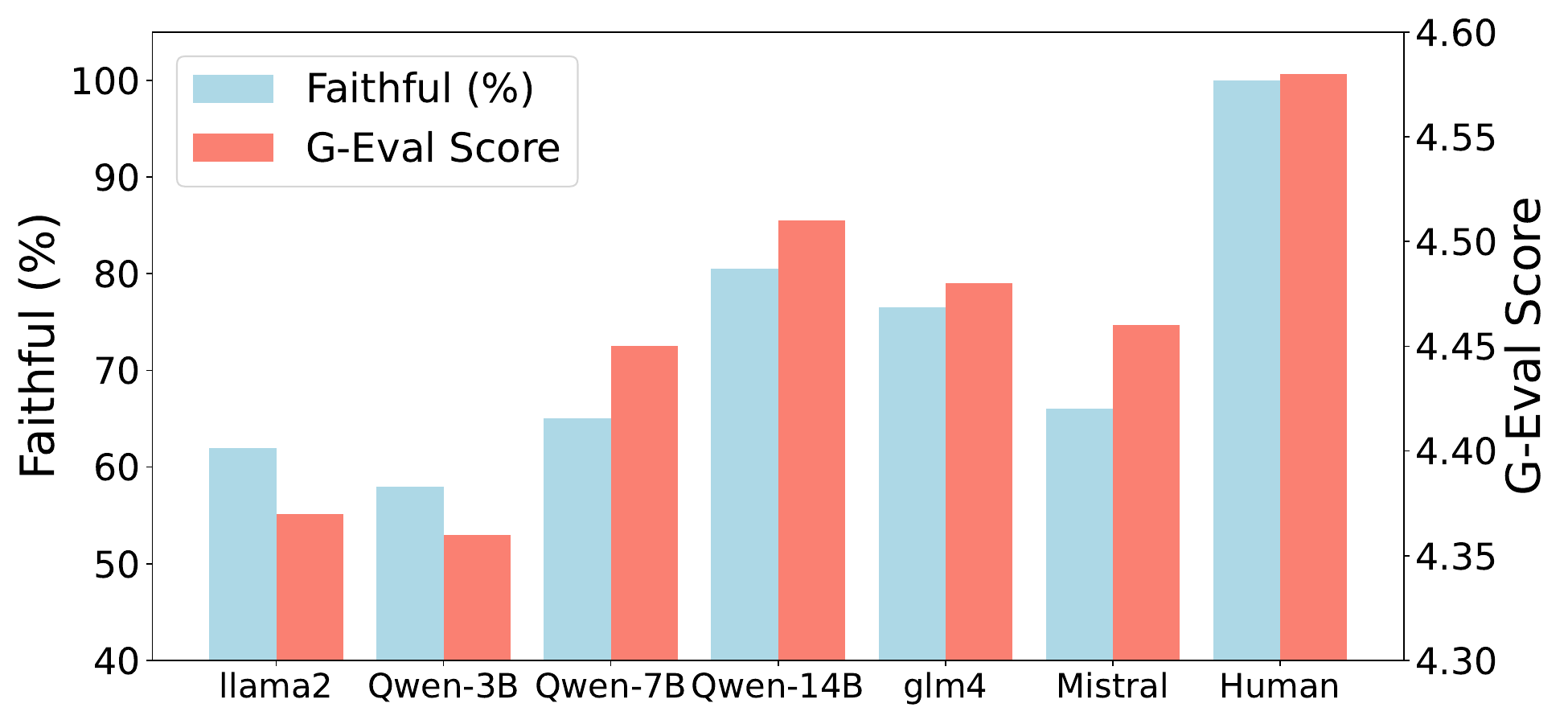}}
  \caption{Pilot experiment.}
  \label{pilot-faith}
\end{figure}

\noindent\textbf{\textsc{FaithDial}} is a benchmark for hallucination-free dialogues, which optimizes the responses in the WoW dataset to be more faithful to knowledge. Subjective and hallucinated information present in the wizard’s utterance of WoW data are edited into utterances faithful to the given knowledge in this dataset. We evaluated all the decoding methods on its test set, which contains 3,539 samples \cite{dziri-etal-2022-faithdial}.

\noindent\textbf{{HalluDial}} is the first comprehensive large-scale benchmark for dialogue-level hallucination evaluation. It is derived from an information-seeking dialogue dataset and covers factuality and faithfulness hallucinations. The benchmark includes 4,094 dialogues with a total of 146,856 samples. We selected 3,000 samples from its 18,357 non-hallucinatory samples as the test set in our experiments.

\section{Baselines} \label{baseline}
\textbf{Beam search} selected the most probable k tokens from the probability distribution at each step to expand the search space \cite{BoulangerLewandowski2013AudioCR, 10.5555/2969033.2969173}. We set the beam size to 4 in our experiment. \\
 \textbf{CS:} \citet{su2022contrastive} penalized previously generated tokens to overcome degeneration and enhance content diversity. We set $k|\alpha=4|0.6$. \\
\textbf{FECS:} \citet{chen-etal-2023-fidelity} extended Contrastive Search by integrating a faithfulness term that encourages factuality. We set $k|\alpha|\beta=4|0.3|0.3$. \\
\textbf{Top-k Sampling} introduced randomness into the generation process by selecting from the top-k most likely tokens \cite{fan-etal-2018-topk}. We set $k=50$ in our experiment. \\
\textbf{Nucleus sampling} considered a dynamic number of words that cumulatively reach the probability $p$ \cite{holtzman2020curious}. We set $p=0.9$ in our experiment. \\
\textbf{F-Nucleus:} \citet{lee2023factuality} modified Nucleus Sampling by adapting the randomness dynamically to improve the factuality of generation. We set $p|\lambda|\omega=0.9|0.9|0.7$ in our experiment. \\
 \textbf{CD:} \citet{li-etal-2023-contrastive} maximized the difference between expert log-probabilities and amateur log-probabilities to improve fluency and diversity. We use Llama2-7B-chat as the expert model and Sheared-LLaMA-2.7B-ShareGPT as the amateur model \cite{xia2023sheared}. We set the amateur temperature $\tau$ to $1.0$. We select the generated tokens using a greedy search on the contrasted distributions.   \\
 \textbf{DoLa:} \citet{chuang2023dola} amplified the factual knowledge in LLM by contrasting the logits from different layers to enhance factuality. We set the $dola\_layers$ hyperparameter to 'high' in our experiments. We select the generated tokens using a greedy search on the contrasted distributions.\\
 \textbf{CAD:} \citet{chuang2023dola} amplified the difference between output probabilities with and without the context document to highlight the external knowledge. We set the hyperparameter $\alpha$ to $1.0$. We select the generated tokens using a greedy search on the contrasted distributions. \\

\section{Implementation Details} \label{imp}
We conducted experiments by utilizing the open-source Hugging Face transformers \cite{transformers}. All experiments are conducted with few-shot prompting (three shots). The three demonstrations are manually selected from the \textsc{FaithDial} dataset, and they are used consistently for all methods during evaluation. We conducted three experiments for all methods, using a different set of samples in each experiment, and finally took the average of the results to eliminate the impact of randomness. We exhibits a set of demonstrations with task instructions in the Appendix \ref{prompt}. As our focus is on the generation process following the acquisition of retrieval knowledge, we assume that the knowledge provided to the model is the most appropriate. For our CoDe and all baselines, we directly use manually annotated golden knowledge from the three datasets as input in the experiments. For the hyperparameters in CoDe, we set $k|\beta|\gamma=4|0.6|3$. For other decoding hyperparameters, we set them to be the same for all methods. We set the $min\_new\_tokens$ to 5, and the $batch\_size$ to 1.

\begin{table}[htbp]
\centering
\begin{tabular}{|ccc|}
\hline
\multicolumn{3}{|c|}{Faithfulness} \\ \hline
\multicolumn{1}{|c|}{BERT-Precision} & \multicolumn{1}{c|}{FaithCritic} & Hallujudge \\ \hline
\multicolumn{3}{|c|}{Expressiveness} \\ \hline
\multicolumn{1}{|c|}{Diversity} & \multicolumn{1}{c|}{Coherence} & Creativeness \\ \hline
\multicolumn{3}{|c|}{Quality} \\ \hline
\multicolumn{1}{|c|}{BLEU} & \multicolumn{1}{c|}{METEOR} & ROUGE-L \\ \hline
\end{tabular}
\caption{Automatic Evaluation metircs.}
\end{table}

\section{Evaluation Metrics} \label{metric}

\subsection{Automatic Evaluation Metrics}

\subsubsection{Faithfulness} To evaluate faithfulness, we adopted three faithfulness evaluation metrics, which has been demonstrated to achieve high correlations with human judgment.

\noindent\textbf{K-BP.} We calculated BERT-Precision \cite{pagnoni-etal-2021-understanding-bertscoreP} between the external knowledge and generated response (\textbf{K-BP}) following \citet{shi-etal-2024-trustingCAD} to measure the consistency from the perspective of semantic similarity.

\noindent\textbf{F-Critic.} FaithCritic is a faithfulness discrimination model fine-tuned on the \textbf{\textsc{FaithCritic}} dataset, which is initialized with RoBERTa-Large. This model outputs the probability of positive and negative labels in the form of a binary-classification Natural Language Inference (NLI) task, where responses with subjective and hallucinatory information are predicted as negative labels. \textbf{F-Critic.} is the average entailment score on the FaithCritic model.

\noindent\textbf{H-Judge.} \textbf{H-Judge} is the ratio of samples judged to be faithful by the Hallujudge model \cite{luo2024halludiallargescalebenchmarkautomatic}. Since the authors did not release the weights of Hallujudge, we trained the Hallujudge model on the HalluDial dataset using Meta-Llama-3-8B with the hyperparameters specified in the paper.

\subsubsection{Expressiveness}

We considered the model's expressiveness in terms of three aspects: diversity, context coherence, and the creativity in knowledge utilization. 

\noindent\textbf{DIV.} Diversity (\textbf{DIV}) is calculated as the geometric mean of Distinct-n (n=1, 2, 3, 4) \cite{li-etal-2016-diversity}:
\begin{equation}
\textbf{DIV}=\sqrt[4]{\prod\limits_{n=1}^{4} \text{Distinct-n}}.
\end{equation}

\noindent\textbf{COH.}
Following \citet{su2022contrastive} and \citet{li-etal-2023-contrastive}, we approximated coherence by cosine similarity between the sentence embeddings of context $x$ and generation $y$:
\begin{equation}
\textbf{COH}=\frac{\text{EMB}(x)\cdot \text{EMB}(y)}{||\text{EMB}(x)||\cdot ||\text{EMB}(y)||}, 
\end{equation}
where EMB($\cdot$) is the SimCSE sentence embedding \cite{gao-etal-2021-simcse}.

\noindent\textbf{CRE.}
To calculated \textbf{CRE}, we use the \textsc{Coverage} divided by the square root of \textsc{Density} \cite{grusky2020newsroom} as follows:
\begin{equation}
\textbf{CRE}=\frac{{\text{Coverage}}}{\sqrt[2]{\text{Density}}}, 
\end{equation}
\begin{equation}
\text{Coverage}(k, y) = \frac{1}{|y|} \sum_{f \in \mathcal{F}(k, y)} |f|,
\end{equation}
\begin{equation}
\text{Density}(k, y) = \frac{1}{|y|} \sum_{f \in \mathcal{F}(k, y)} |f|^2,
\end{equation}
where $\mathcal{F}(k, y)$ is the set of shared sequences of tokens in knowledge $k$ and response $y$. A higher Coverage score indicates more knowledge are integrated into the response, while a lower Density score indicates the knowledge are weaved into response naturally and creatively. To unify the measurement of Coverage and Density, we performed a square root operation on Density.

\subsubsection{Quality}
To assess the overall quality of generated responses, we selected three widely-used metrics based on calculating overlap with the ground-truth: \textbf{BLEU} \cite{2002-bleu}, \textbf{METEOR} \cite{banerjee-lavie-2005-meteor}, and \textbf{ROUGE} \cite{lin-2004-rouge}.

\subsection{LLM-based Evaluation Metrics} \label{llm-eval}

We perform LLM-based evaluation according to the following criteria: Naturalness (\textbf{Nat.}), Coherence (\textbf{Coh.}), Informativeness (\textbf{Inf.}), Creativity (\textbf{Cre.}), Faithfulness (\textbf{Fai.}), and Factuality (\textbf{Fac.}). 
High naturalness refers to the generated content being realistic, engaging, and interactive, capable of encouraging users to engage in more rounds of conversation.
High coherence means the generated content is related to the context and maintains a smooth flow.
High informativeness indicates that the generated content is rich in information and can help users acquire new knowledge.
High creativity refers to the model's diverse utilization of external knowledge, rather than mechanically extracting and directly outputting it.
High faithfulness indicates that the generated content does not contain information that directly contradicts the given knowledge, or cannot be verified from the provided knowledge.
High factuality indicates that the generated content does not conflict with established world knowledge. To strictly differentiate between faithfulness and factuality, we included detailed definitions for both in Appendix \ref{faithfact}. Table \ref{tab:gpt4o} shows the prompts we used for LLM-based evaluation (LLM-as-a-Judge). Our evaluation was conducted using gpt-4o-2024-08-06.

\subsection{Human Evaluation Metrics}

\begin{figure*}[htbp]
  \centerline{\includegraphics[scale=0.55]{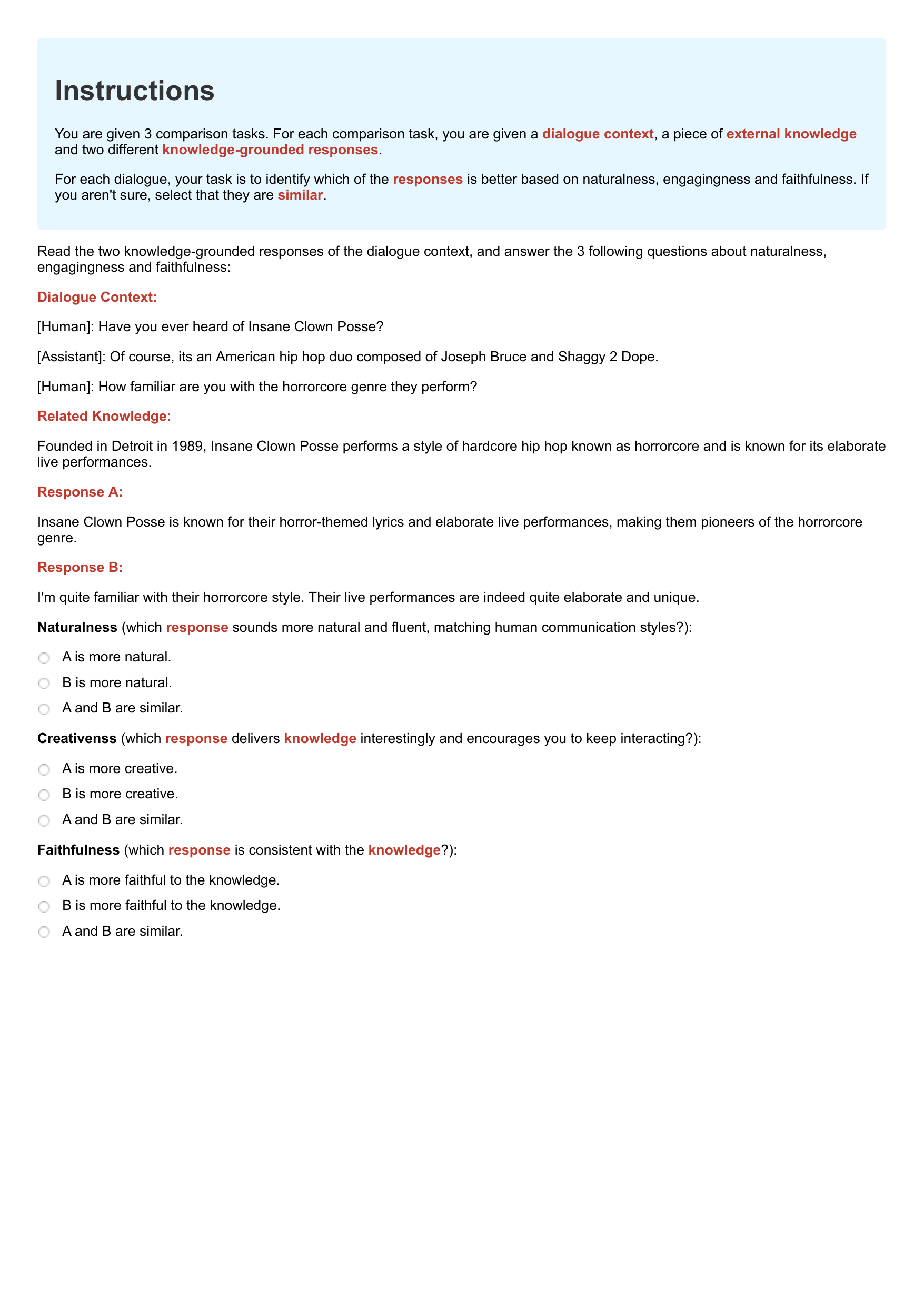}}
  \caption{Human evaluation instructions and interface.}
  \label{fighuman-1}
\end{figure*}

Given the dialogue context, related knowledge and the responses generated by CoDe and its baselines, five well-educated annotators were asked to choose the superior response based on three criteria:  Naturalness (\textbf{Nat.}), Engagingness (\textbf{Eng.}), and Faithfulness (\textbf{Fai.}). Detailed evaluation guidelines can be seen in Figure \ref{fighuman-1}.

\section{Additional Related Work}

\subsection{Knowledge-Grounded Dialog Generation}
Knowledge-grounded dialogue generation aims to alleviate dull and unfaithful responses by injecting external knowledge into input of dialogue models and it consists of two subtasks: knowledge selection and response generation. The hot spot of early research is mainly concentrated on how to improve the performance of knowledge selection \cite{eacl-2023-generative,xu-etal-2022-corefdiffs, 2021-colv,yang-etal-2022-take,MIKE,kim2020SKT,Wang2022seek,wang-etal-2025-sibyl,yang-etal-2023-multi-level}. With the remarkable leap in the capabilities of generative models, the research focus gradually shifts to the response generation subtask \cite{zhao2020lowresource,liu2021threestage,zhao2020knowledgegrounded,zheng-etal-2021-knowledge,chen2024multitaskroleplayingagentcapable,yang2025dynamicearlyexitreasoning}. Ideally, a brilliant robot should generate informative and truthful responses while maintaining the naturalistic phrasing and excellent interactivity \cite{dziri-etal-2022-faithdial}.

\section{Details of the Generation Prompts}
\label{prompt}
Our instruction template and demonstrations for prompting Large Language Models to generate response during evalutaion are as follows: 

\textit{As an assistant, your task is to engage in a chit-chat conversation with user. You will be provided the dialogue history and a piece of related knowledge, and your task is to utilize the knowledge to continue the conversation. Your English response should be informative but no more than 50 words, coherent with the dialogue context and faithful to the given knowledge. \\ You SHOULD refer to the following examples: \\
Example 1: \\
User's utterance: Shower. \\
\#\#\#knowledge\#\#\#: Ancient people began to reproduce these natural phenomena by pouring jugs of water, often very cold, over themselves after washing.\\
Your knowledge-grounded response to the user: I love taking showers! I do not know how I would live without the modern showers. In ancient times people would pour jugs of cold water over themselves and consider that a shower. \\
Example 2: \\
User's utterance: Is rock and roll still popular today? \\
Your response: It's hard to say. However, radio stations have much success playing classic rock and roll, which is a sub genre that usually has one or two electric guitars, a double bass or string bass or electric bass guitar, and a drum kit. \\
User's utterance: I used to listen to the rock band Rolling Stones.  Are they still around today? \\
Your response: They are! Even though they were formed in 1962 and have had a long list of line-up changes, they're still around today, with Mick Jagger still leading the band. \\
User's utterance: Wow, that is a long time to be playing music. I wonder if any other bands have been around that long.\\
\#\#\#knowledge\#\#\#: Red Hot Chili Peppers are an American funk rock band formed in Los Angeles in 1983. \\
Your knowledge-grounded response to the user: It all depends! You have bands like the Red Hot Chili Peppers who, although have not reached the popularity of the Rolling Stones, have been around since 1983 themselves.
\\
Example 3: \\
User's utterance: jazz music is a very interesting sound and interesting genre. what can you tell me about it? \\
\#\#\#knowledge\#\#\#: Jazz is a music genre that originated in the African-American communities of New Orleans, United States, in the late 19th and early 20th centuries, and developed from roots in blues and ragtime. \\
Your knowledge-grounded response to the user: I like jazz too. It originated in the African-American communities of New Orleans in the late 19th and 20th centuries. \\
Now complete the following dialogue:\\
User's utterance: [Dialogue Context]\\
\#\#\#knowledge\#\#\#: [External Knowledge]\\
Your knowledge-grounded response to the user: }








\begin{table*}[]
\centering
\scalebox{0.9}
{
\begin{tabular}{ll}
\bottomrule

\textbf{LLM-as-a-Judge}  &  
\begin{tabular}[c]{@{}l@{}}
Please act as an impartial judge and evaluate the quality of the response provided\\ by an AI assistant to the user utterance displayed below. You will be provided with\\ the related knowledge, a dialogue history, and a generated response. \\
Your evaluation should be mainly based on $<$evaluation criteria$>$ \\
Begin your evaluation by providing a short explanation. Be as objective as possible.\\ After providing your explanation, you must rate the response on a scale of 1 to 5.\\
The rating is: \\
\end{tabular}        
 \\ \hline
\bottomrule
\end{tabular}
}
\centering
\caption{The prompts for LLM-based evaluations.}
\label{tab:gpt4o}
\end{table*}

\begin{table*}[]
\centering
\scalebox{0.9}
{
\begin{tabular}{ll}
\bottomrule

\textbf{Context} & 
\begin{tabular}[c]{@{}l@{}}
 $<$Human$>$ Madonna is one of my favorite entertainers, do you know much about her? \\
 $<$Assistant$>$ Sure! She was born in 1958 on August 16, and is an artist and an actress. \\
 $<$Human$>$ I didn't know she was that old, it doesn't matter though she's such \\ a great artist anyway! \\ 
 $<$Assistant$>$ I see, well she is also cited by other artists as an influence. \\
 $<$Human$>$ I certainly think she's an influence as well, do you know any of Madonna's \\ most famous songs?
\end{tabular}  \\  

\hline
\textbf{Knowledge} &   
\begin{tabular}[c]{@{}l@{}}
{\color[HTML]{3166FF} She followed it with a series of successful albums, including the global bestsellers,}\\
{\color[HTML]{3166FF}''Like a Virgin'' (1984) and ''True Blue'' (1986), as well as the Grammy Award winners,}\\
{\color[HTML]{3166FF}''Ray of Light'' (1998) and ''Confessions on a Dance Floor'' (2005).}
\end{tabular}           
\\\hline
\textbf{Greedy}  &  
\begin{tabular}[c]{@{}l@{}}
Yes! Some of her most famous {\color[HTML]{FE0000}songs} are {\color[HTML]{FE0000}Material Girl, American Pie}, {\color[HTML]{3166FF}Like a Virgin,}\\ and {\color[HTML]{3166FF}Ray of Light}.
\end{tabular}     \\\hline

\textbf{Nucleus}  &  
\begin{tabular}[c]{@{}l@{}}
Some of her most famous {\color[HTML]{FE0000}songs} include {\color[HTML]{3166FF}Like a Virgin}, {\color[HTML]{FE0000}Material Girl, American Pie,} \\and {\color[HTML]{FE0000}Hung Up}.
\end{tabular}        
 \\\hline
\textbf{DoLa}  &  
\begin{tabular}[c]{@{}l@{}}
 Of course! Some of her most famous {\color[HTML]{FE0000}songs} include {\color[HTML]{3166FF}"Like a Virgin}," {\color[HTML]{FE0000}"Material Girl,"}\\ {\color[HTML]{FE0000}"Vogue," "Into the Groove," "Like a Prayer," "Express Yourself," "Cherish," "Frozen,"}
 \\{\color[HTML]{FE0000}"Hung Up," "4 Minutes"}...(More generated songs omitted here)
\end{tabular}        
 \\\hline
 \textbf{CAD}  &
 \begin{tabular}[c]{@{}l@{}}
 She has a series of successful albums,{\color[HTML]{3166FF} including the global bestsellers,}\\
{\color[HTML]{3166FF}''Like a Virgin'' (1984) and ''True Blue'' (1986), as well as the Grammy Award winners,}\\
{\color[HTML]{3166FF}''Ray of Light'' (1998) and ''Confessions on a Dance Floor'' (2005).}
\end{tabular}          
\\\hline
\textbf{CoDe}  &
\begin{tabular}[c]{@{}l@{}} Of course! She's had many! {\color[HTML]{3166FF}Like a Virgin, True Blue, Ray of Ligh,} {\color[HTML]{3166FF}Confessions on a}\\ {\color[HTML]{3166FF} Dance Floor,}  are some of her most famous albums. Which song do you like the most? \\
\end{tabular}          
\\
\bottomrule
\end{tabular}
}
\centering
\caption{An example for case study. The song "American Pie" was covered by Madonna. The baseline confused song titles with album names and generated many songs that cannot be verified by external knowledge.}
\label{tab:case1}
\end{table*}

\begin{table*}[]
\centering
\scalebox{0.93}
{
\begin{tabular}{ll}
\bottomrule

\textbf{Context} & 
\begin{tabular}[c]{@{}l@{}}
 $<$Human$>$ Are you a basketball fan? \\
 $<$Assistant$>$ Yes, I am a fan of the five player sport. Are you?\\
 $<$Human$>$ Not as much as I used to be. I watched The Bulls in the 90s when they were\\ the Dream Team. \\ 
\end{tabular}  \\  

\hline
\textbf{Knowledge} &   
\begin{tabular}[c]{@{}l@{}}
{\color[HTML]{3166FF}Jordan played 15 seasons in the NBA for the Chicago Bulls and Washington Wizards.}
\end{tabular}    
\\ \hline
\textbf{Beam}  &  
\begin{tabular}[c]{@{}l@{}}
Interesting!  {\color[HTML]{3166FF} Michael Jordan played 15 seasons in the NBA for the Chicago Bulls and}\\ {\color[HTML]{3166FF}Washington Wizards.}
\end{tabular}     \\ \hline

\textbf{Nucleus}  &  
\begin{tabular}[c]{@{}l@{}}
Wow, impressive! Even though Michael Jordan {\color[HTML]{FE0000} played for several teams}, he is mostly\\ known for { \color[HTML]{FE0000} his time with the Chicago Bulls on the Dream Team}.
\end{tabular}        
 \\ \hline
\textbf{CAD}  &  
\begin{tabular}[c]{@{}l@{}}
 I see. {\color[HTML]{3166FF} Jordan played 15 seasons in the NBA for the Chicago Bulls and}\\ {\color[HTML]{3166FF}Washington Wizards.}
\end{tabular}        
 \\ \hline
\textbf{CoDe}  &
\begin{tabular}[c]{@{}l@{}}  Wow, that's amazing! Michael Jordan played for Chicago Bulls and Washington  Wizards \\ for 15 seasons. Do you like the  Bulls because of Michael Jordan?
  \\
\end{tabular}          
\\
\bottomrule
\end{tabular}
}
\centering
\caption{An example for case study. Jordan only played for two teams, so it is inaccurate for Nucleus to say that he played for several teams. The "Dream Team" generally refers to the U.S. Olympic men's basketball team and should be distinguished from NBA games.}
\label{tab:case2}
\end{table*}

\begin{table*}[]
\centering
\scalebox{0.9}
{
\begin{tabular}{ll}
\bottomrule

\textbf{Context} & 
\begin{tabular}[c]{@{}l@{}}
 $<$Human$>$ Hi, I like to volunteer at a nursing home, helping the elderly! \\
\end{tabular}  \\  

\hline
\textbf{Knowledge} &   
\begin{tabular}[c]{@{}l@{}}
{\color[HTML]{3166FF}Nursing homes will provide short-term rehabilitative stays following a surgery, illness}\\ {\color[HTML]{3166FF}or injury which may require physical therapy, occupational therapy or speech-language} \\ {\color[HTML]{3166FF}therapy}.
\end{tabular}    
\\ \hline
\textbf{Greedy}  &  
\begin{tabular}[c]{@{}l@{}}
That's great to hear that you volunteer at a nursing home. Many nursing homes also\\ offer {\color[HTML]{FE0000}long-term care} for residents who need ongoing assistance with daily activities.
\end{tabular}     \\\hline

\textbf{Nucleus}  &  
\begin{tabular}[c]{@{}l@{}}
Great to hear that you're interested in volunteering at a nursing home!
\end{tabular}        
 \\\hline
\textbf{CAD}  &  
\begin{tabular}[c]{@{}l@{}}
Volunteering at a nursing home is a wonderful way to support the elderly.
Many \\
{\color[HTML]{3166FF}nursing homes offer short-term rehabilitative stays following surgeries, illnesses, or injuries,}\\ {\color[HTML]{3166FF}which may require physical therapy, occupational therapy or speech-language therapy.}
\end{tabular}        
 \\\hline
\textbf{CoDe}  &
\begin{tabular}[c]{@{}l@{}}  That's great! You can assist those who need {\color[HTML]{3166FF}physical,  occupational, or speech therapy} \\during their recovery. What a great job!
  \\
\end{tabular}          
\\
\bottomrule
\end{tabular}
}
\centering
\caption{An example for case study. }
\label{tab:case3}
\end{table*}

\section{Faithfulness and Factuality} \label{faithfact}

We formally define faithfulness and factuality and explain their relationship to facilitate the research:
 
\noindent\textbf{Definition 1} Faithfulness($\mathcal{F}$):
\textit{Given a response} $y$, \textit{and external knowledge} $\mathcal{K} = (k_1, \ldots, k_j)$ \textit{at turn} $n$, \textit{we say that} \textit{the response $y$ is faithful with respect to the external knowledge} ($\mathcal{F}(\mathcal{K},y)$) \textit{if and only if the following condition holds}:
\begin{itemize}
    \item  $\exists\, \Gamma$ \textit{such that} $\Gamma \models y$, \textit{where} $\Gamma$ \textit{is a non-empty subset of} $\mathcal{K}$ \textit{and} $\models$ \textit{denotes semantic entailment}. \textit{In other words, there is no interpretation} $\mathcal{I}$ \textit{such that all members of} $\Gamma$ \textit{are true and} $y$ \textit{is false} \cite{dziri-etal-2022-faithdial}.
\end{itemize}
\noindent\textbf{Definition 2} Factuality($\mathcal{T}$):
\textit{Given a response} $y$, \textit{we say that} $y$ \textit{is factual} ($\mathcal{T}(y)$) \textit{if and only if the following condition holds}:
\begin{itemize}
    \item  $\exists\, \Phi$ \textit{such that} $\Phi \models y$, \textit{where}  $\Phi$ \textit{is a non-empty subset of world knowledge} $\mathcal{K}_w$ \textit{and} $\models$ \textit{denotes semantic entailment}.
\end{itemize}
\noindent\textbf{Theorem 1} \label{theorem} $\mathcal{F} \models \mathcal{T}, \mathcal{T} \not\models \mathcal{F}$, where $\models$ denotes entailment.

Theorem 1 indicates that responses ensuring faithfulness are necessarily factual, but the converse may not always apply. The proof of the theorem is shown in the Appendix \ref{proof}.

\section{The proof of Theorem 1}
\label{proof}

\noindent\textbf{Theorem.} $\mathcal{F} \models \mathcal{T}, \mathcal{T} \not\models \mathcal{F}$.

\begin{proof}
$\mathcal{F} \models \mathcal{T}:$
For all $y$ that satisfy $\mathcal{F}(K,y)$, there exists $\Gamma$ such that $\Gamma \models y$ and $\Gamma \subseteq K$. Since $K \subsetneq K_{w}$ (external knowledge is a proper subset of world knowledge), it follows that $\Gamma \subseteq K_{w}$. Let $\Phi=\Gamma$, then  $\Phi \models y$ and $\Phi \subseteq K$. Hence, $\mathcal{T}(y)$ holds, and the conclusion is proved. 

$\mathcal{T} \not\models \mathcal{F}:$
We prove it by contradiction. Suppose that $\mathcal{T} \models \mathcal{F}$, then for all 
$y$ that satisfy $\mathcal{T}(y)$, there exists $\Phi$ such that $\Phi \models y$ and $\Phi \subseteq K_w$. Let $\Phi \subseteq K_w / K$. Since $\mathcal{T} \models \mathcal{F}$, then $\Phi \subseteq K$. However, $\Phi \subseteq K_w / K$, it implies that $\Phi = \emptyset$, but $\Phi \neq \emptyset$, leading to a contradiction, thus the conclusion is not valid.

\end{proof}

\end{document}